\documentclass[times,twocolumn,final]{elsarticle}

\usepackage{framed,multirow}
\usepackage{float}

\usepackage{amssymb}
\usepackage{latexsym}
\usepackage{amsmath}

\usepackage{url}
\usepackage{xcolor}
\usepackage{placeins}
\usepackage{hyperref}

\definecolor{newcolor}{rgb}{.8,.349,.1}

\journal{Medical Image Analysis}

\begin{document}


\begin{frontmatter}

\title{\LARGE SlideGraph$^{+}$: Whole Slide Image Level Graphs to Predict HER2 Status in Breast Cancer}

\author[1]{Wenqi Lu}
\author[2]{Michael Toss}
\author[2]{Emad Rakha}
\author[1]{Nasir Rajpoot}
\author[1]{Fayyaz Minhas\corref{cor1}}
\cortext[cor1]{Corresponding author: 
  Tel.: +44 (24) 7652 4314;}
\ead{Fayyaz.Minhas@warwick.ac.uk}

\address[1]{Tissue Image Analytics (TIA) Centre, Department of Computer Science, University of Warwick, UK}
\address[2]{Nottingham Breast Cancer Research Centre, Division of Cancer and Stem Cells, \\School of Medicine, Nottingham City Hospital, University of Nottingham, Nottingham, UK}


\begin{abstract}
Human epidermal growth factor receptor 2 (HER2) is an important prognostic and predictive factor which is overexpressed in 15-20\% of breast cancer (BCa). The determination of its status is a key clinical decision making step for selection of treatment regimen and prognostication. HER2 status is evaluated using transcroptomics or immunohistochemistry (IHC) through situ hybridisation (ISH) which require additional costs and tissue burden in addition to analytical variabilities in terms of manual observational biases in scoring. In this study, we propose a novel graph neural network (GNN) based model (termed SlideGraph$^{+}$) to predict HER2 status directly from whole-slide images of routine Haematoxylin and Eosin (H\&E) slides. The network was trained and tested on slides from The Cancer Genome Atlas (TCGA) in addition to two independent test datasets. We demonstrate that the proposed model outperforms the state-of-the-art methods with area under the ROC curve (AUC) values $>$ 0.75 on TCGA and 0.8 on independent test sets. Our experiments show that the proposed approach can be utilised for case triaging as well as pre-ordering diagnostic tests in a diagnostic setting. It can also be used for other weakly supervised prediction problems in computational pathology. The SlideGraph$^{+}$ code is available at  \url{https://github.com/wenqi006/SlideGraph}. 
\end{abstract}


\end{frontmatter}


\section{Introduction}

Breast cancer (BCa) is the most commonly diagnosed cancer among women, and the second leading cause of female cancer related deaths worldwide \citep{ahmad2019breast}. Human epidermal growth factor receptor 2 (HER2) positivity accounts for around 15\% of the early stage BCa. HER2 positivity in BCa is defined as evidence of HER2 protein overexpression and/or HER2 gene amplification \citep{ross2009her} which is proved to be associated with worse clinical outcome \citep{slamon1987human}. HER2-positive BCa tumours tend to grow and spread faster than HER2-negative tumours, but are much more likely to respond to targeted therapy with anti-HER2 drugs \citep{yarden2001biology,nahta2006mechanisms}.

\begin{figure*}[h!]
\centering
\includegraphics[width=1\textwidth]{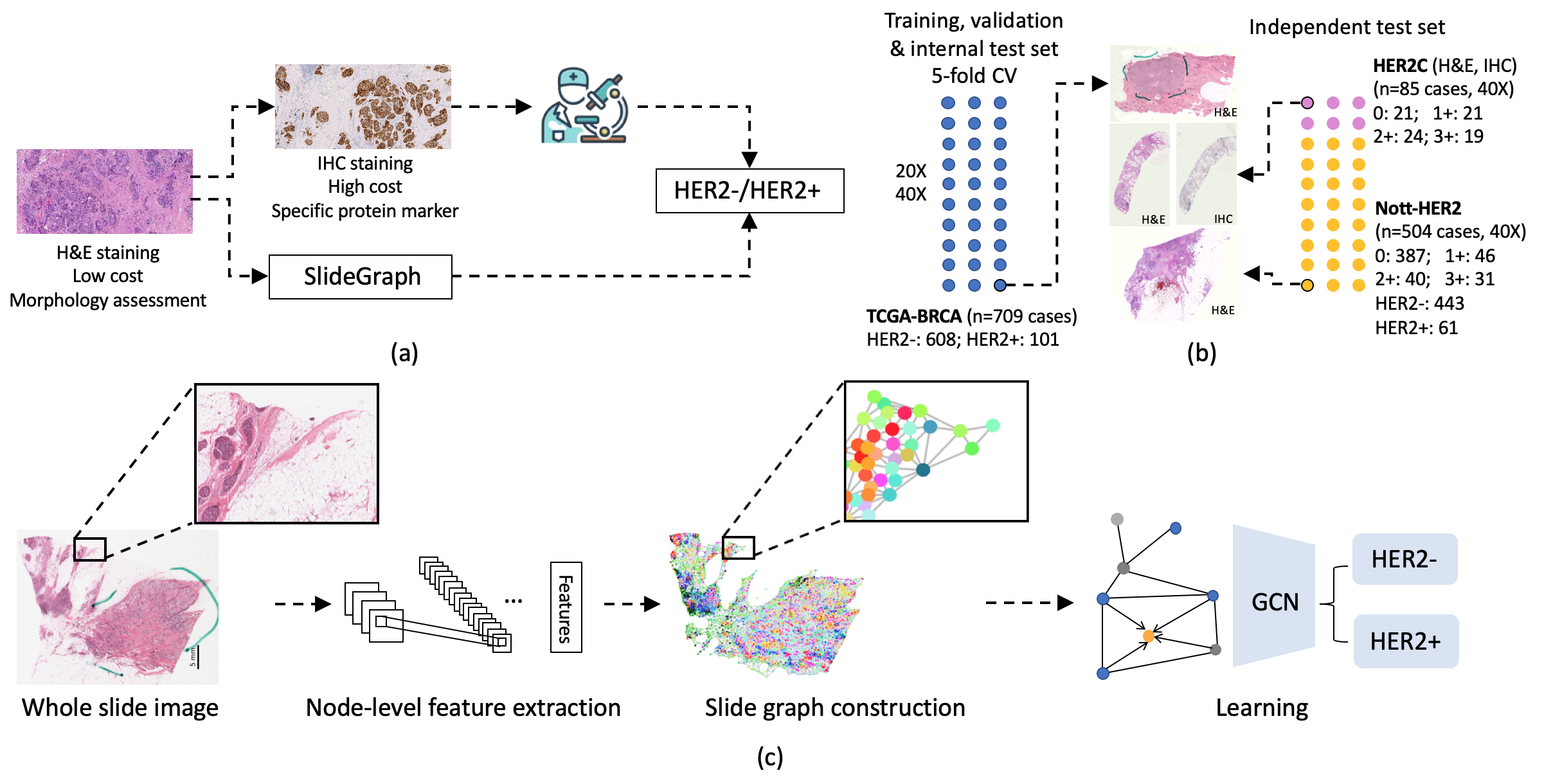}
\caption{HER2 status prediction from H\&E images: (a) In routine diagnostic practice of BCa, tissue sections are commonly stained with H\&E, followed by immunohistochemistry (IHC) staining to estimate the presence of specific protein receptors. We show that our deep learning algorithm can predict HER2 directly from H\&E images; (b) Multi-centre datasets that are used to train, validate and test our proposed model; (c) Without pixel-level annotations, our algorithm is trained on the WSI-level graphs and predicts HER2 status directly.
} \label{overflow}
\end{figure*}

In routine diagnostic practice, BCa tissue sections are stained with Haematoxylin and Eosin (H\&E) and visually examined for morphological assessment. It is then followed by ancillary techniques including immunohistochemistry (IHC) and in situ hybridisation (ISH) to assess the expression of specific proteins, including HER2, for prognostic and predictive purposes (Fig.~\ref{overflow}(a)). The current guidelines \citep{wolff2018human} revised by the American Society of Clinical Oncology/College of American Pathologists (ASCO/CAP) assign a HER2 positivity score between 0 and 3+ based on visual analysis of IHC slides. Cases scoring 0 or 1+ are classified as HER2-negative (HER2-), while cases with a score of 3+ are regarded as HER2-positive (HER2+). Cases with score 2+ refer to equivocal expression of HER2 that need further assessment using ISH to evaluate HER2 gene status. Operational and analytical limitations of aforementioned techniques in terms of cost, tissue usability  and observer-subjectivity in manual scoring affect interpretation of HER2 status and hence patient management. Consequently, prediction of HER2 status directly from digitally scanned whole slide images (WSIs) of routine H\&E-stained tissue sections through deep learning or Artificial Intelligence (AI) techniques is of significant clinical and scientific interest. 

\begin{figure}[h!]
\begin{center}
\includegraphics[width=0.5
\textwidth]{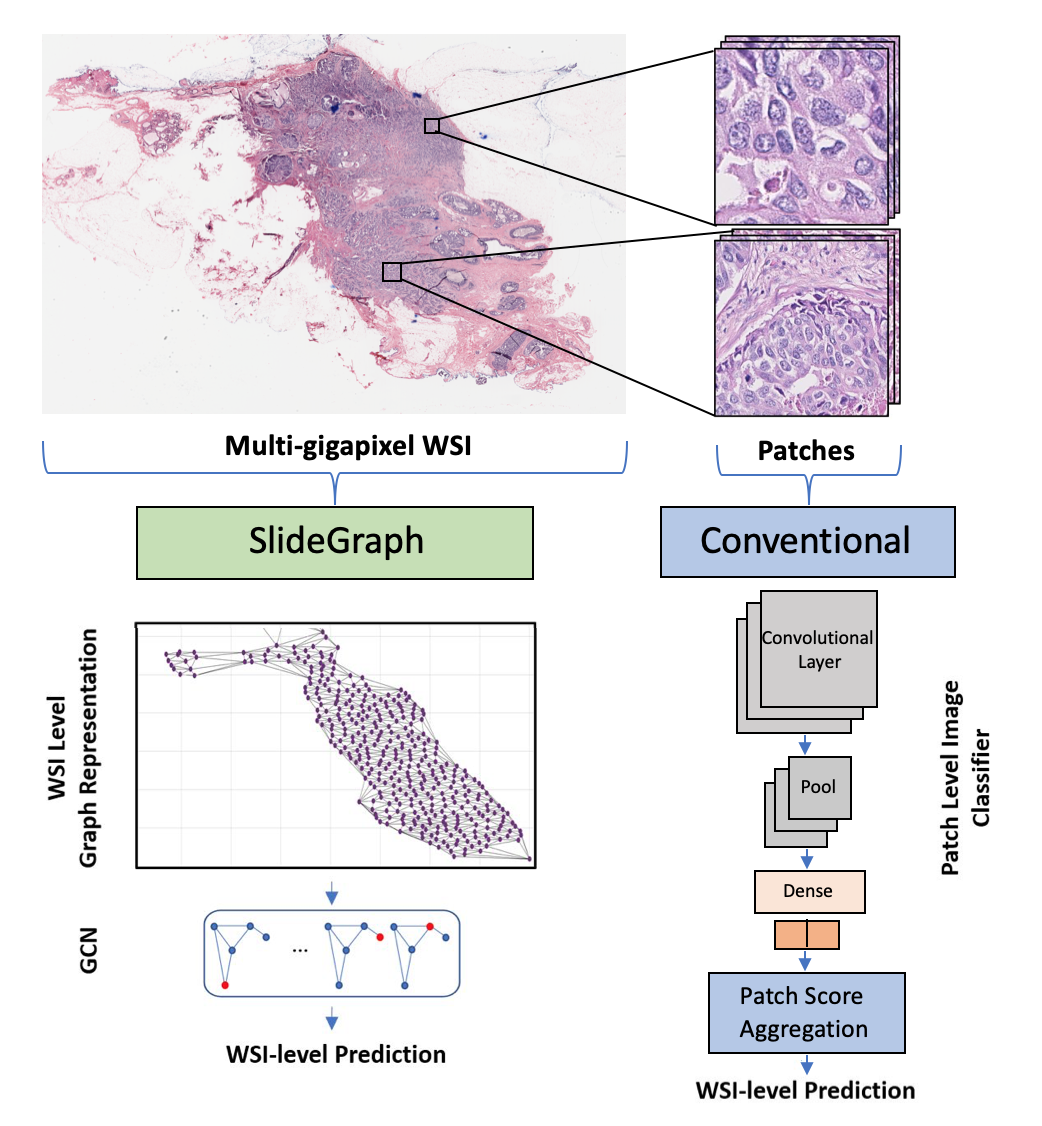}
\end{center}
\caption{The proposed SlideGraph model vs. conventional patch-based methods. Our proposed SlideGraph model is able to capture the overall organisation and structure of the tissue.} \label{patchlevel}
\end{figure}

Digital pathology and AI offer significant potential to overcome the aforementioned limitations and improve reproducibility (\cite{qaiser2018her, acs2020artificial, farahmand2021deep}). Such computational pathology (CPath) models have been used for diagnostics as well as prediction of genetic expression correlates. Kather \textit{et al.} \citep{kather2019pan} proposed a deep learning method to predict hormone receptor status from routine H\&E WSIs. Morphological correlates of specific mutations have also been observed in H\&E stained BCa histology images. Rawat \textit{et al.} \citep{rawat2020deep} introduced the concept of ``tissue fingerprints'' to learn H\&E features that can distinguish one patient from another. However, a major limitation of existing AI methods stems from patch-level analysis employed by these methods. As an entire WSI at full-resolution can be of the order of 150,000$\times$100,000 pixels, training a model on the full-resolution WSIs is computationally challenging and expensive. A two-step patch-level approach is typically used to deal with large size WSIs (Fig.~\ref{patchlevel}) \citep{janowczyk2016deep,bandi2018detection}. First, the image is divided into small image tiles (or patches), where each patch is processed independently by the neural network \citep{tizhoosh2018artificial}. Then predicted scores for each patch within the WSI are aggregated into a WSI-level score, usually by pooling their results with various aggregating strategies such as average pooling, max pooling and majority voting  \citep{cruz2014automatic,lecun1998gradient,nguyen2009weakly}. 

Patch-level analysis used in conventional CPath models has two major drawbacks. First, local patches have a limited visual context. The optimal resolution and patch size for analysis are highly problem-dependent \citep{hou2016patch}. Image patches at a high magnification level lead to loss of contextual information whereas patches at lower magnification levels may not capture cell-level features (Fig.~\ref{patchlevel}). Consequently, a patch-level machine learning method cannot capture the overall organisation and structure of the tissue in a WSI. Second, in most prediction problems in computational pathology, only WSI-level labels are available. It is non-trivial to model the association of each patch with a specific label. Weakly supervised machine learning methods such as Multiple Instance Learning (MIL) have been proposed to alleviate these problems about labels of training patches and aggregate patch-level predictions into WSI-level classification \citep{andrews2002multiple,zhang2006multiple,zhang2002dd,campanella2019clinical}. However, these methods are unable to model the overall organisation and structure of the tissue at both global and local levels. As a consequence, graph-based approaches in computational pathology present a more principled way of modelling such prediction tasks. 

The cell-graph technique \citep{prewitt1979graphs,gunduz2004cell,demir2005augmented,jaume2021histocartography} was introduced to learn the structure-function relationship by modelling geometric structure of the tissue using graph theory. It is based on the assumption that cells in a tissue can organise in a certain way for specific functional states. Such cell-graphs can have different types, such as Delaunay triangles \citep{weyn1999computer,chew1989constrained}, Voronoi diagrams, Minimum Spanning Trees (MST), and Cell Cluster Graphs (CCG) \citep{ali2013cell}. Yener \citep{yener2016cell} explored various cell-graph constructions to establish a quantitative relationship between the geometric structure and functional states. Cell-graph constructions have been successfully used to characterise spatial proximity of histopathological primitives in tasks, such as survival prediction in lung cancer \citep{lu2018feature}, risk stratification in BCa \citep{whitney2018quantitative} and distant metastasis prediction in colorectal cancer \citep{sirinukunwattana2018novel}. However, all these graph-based methods with deep learning classifiers were all trained on image patches which have limited visual context. In addition, since these methods have not been applied on the WSI level, extra patch-based voting methods are necessary to predict the label of a given WSI.

In this study, we propose a graph neural network model, termed as SlideGraph$^{+}$ to address the limitations of existing methods. Instead of extracting small patches from the WSI and doing analysis on a limited visual field for prediction, we introduce a novel pipeline which operates on a graph at the entire WSI-level for prediction of HER2 status. A graph neural network is then used for WSI-level prediction. This work is a significant extension of our previously proposed SlideGraph \citep{lu2020capturing} and employs an extended network architecture that makes the model more interpretable. Outputs from the proposed network not only cover the overall prediction score, but also show the active graph nodes (i.e., image regions) which contribute to the overall prediction. The proposed SlideGraph$^{+}$ model also incorporates a novel message passing technique and a modified loss function. This method accounts for both cell-level information and contextual information by modelling cellular architecture and interactions in the form of a graph. We demonstrate the effectiveness of the proposed scheme on clinically relevant prediction problems from BCa H\&E WSIs. Specifically, we train a classification model to predict the status of HER2 and test it on another two independent cohorts (Fig.~\ref{overflow}(b)). Overall, our main contributions in this paper can be summarised as follows:
 
\begin{itemize}
  \item SlideGraph is the first method which can generate whole slide image level predictions by using a graph representation of the cellular interconnection geometry in a WSI.
  \item The proposed SlideGraph$^{+}$ network architecture is an extension of our previously proposed SlideGraph \citep{lu2020capturing} with an architecture layout which makes the network more interpretable by generating node-level predictions.
  \item SlideGraph$^{+}$ makes use of nuclear composition, cell morphology, neural network embeddings or DAB density estimates features to represent the complex organisation of cells and the overall tissue micro-architecture. The proposed network outperforms the state-of-the-art methods by a significant margin in HER2 status prediction.
  \item The DAB density regression model proposed in this paper is the first method to predict DAB intensity directly from H\&E stained images. It carries potential of removing the necessity of IHC staining when evaluating the HER2 expression.
  \item Instead of annotating invasive tumour regions which is very time-consuming, SlideGraph$^{+}$ is trained on all tissue regions and is able to precisely localise the regions that contribute to the HER2 positivity and expression.
  \item Our trained HER2 status prediction model is tested on two independent cohorts, demonstrating its generalisation on multi-centre datasets.
  \item SlideGraph is computationally more efficient than patch-based models and opens the avenue of using WSI graph representations for solving other problems in computational pathology.

\end{itemize}

\section{Methodology}
The proposed framework for predicting the receptor status from H\&E images is shown in Fig.~\ref{overflow}(c). A typical weakly supervised machine learning problem in computational pathology involves a training dataset $\{(X_i,y_i)|i=1...M\}$ of $M$ WSIs denoted by $X_i$, each with a label $y_i\in\{0, 1\}$. The objective is then to develop a machine learning model that can predict the label for unseen cases. In this work, we build a graph representation  $G_i = G\left ( X_i  \right)$ of each $X_i$ in the training set and train a graph neural network with  trainable parameters $\mathbf{\theta}$ to generate slide-level predictions $F\left( G(X_i);\mathbf{\theta}  \right)$. The trained model $F$ is used for inference to predict status for WSIs which are not included in the training set.

The overall framework consists of four steps: first, we extract features from local regions in the WSI after preprocessing. Specifically, a given WSI $X$  is modelled as a set of image patches $x_{j}\in X$ of size 224$\times$224 pixels at 40$\times$ magnification. Each patch $x_{j}\equiv \left ( \boldsymbol{g}_j, \boldsymbol{h}_j \right )$ is represented as a tuple consisting of a $d$-dimensional feature vector representation $\boldsymbol{h}_j \in \mathbb{R}^d$ and the corresponding geometric coordinates $\boldsymbol{g}_j \in \mathbb{R}^2$ of the top-left corner of the patch. Second, we use spatial clustering to group neighbouring image patches with similar features into clusters. Third, a graph representation based on these clusters is generated to capture the cellular and morphological topology of the WSI. Finally, the graph constructed from the entire WSI is taken as an input to a graph neural network to predict the receptor status at the graph node-level and also at the slide-level. Below, we give details of the datasets and individual steps in the proposed pipeline. 

\subsection{Datasets}
The training dataset used in this study
was obtained from The Cancer Genome Atlas in breast cancer (TCGA-BRCA) \citep{cancer2012comprehensive}. Molecular status of HER2 was assessed clinically on the patient level. We used five-fold stratified cross-validation for a direct comparison with other patch-based classification methods \citep{rawat2020deep,kather2019pan}. In each fold-run, 20\% of the dataset (at the patient level) was held out as unseen test data whereas the remaining 80\% was used for training and validation. We then tested the trained model on other two independent cohorts, the publicly available HER2Contest challenge (HRE2C) dataset \citep{qaiser2018her} and an internal Nottingham University Hospital (Nott-HER2) dataset, respectively. Some high-level information of all three datasets is shown in Fig.~\ref{overflow}(b).

\subsection{Pre-processing}
We use stain normalisation technique by Vahadane \textit{et al.} \citep{vahadane2016structure} to normalise the stain distribution across slides especially those from different centres. Tissue segmentation is performed to remove background regions. The segmented tissue region is then divided into a set of uniformly sized patches.

\begin{figure*}[h]
\begin{center}
\includegraphics[width=1.0\textwidth]{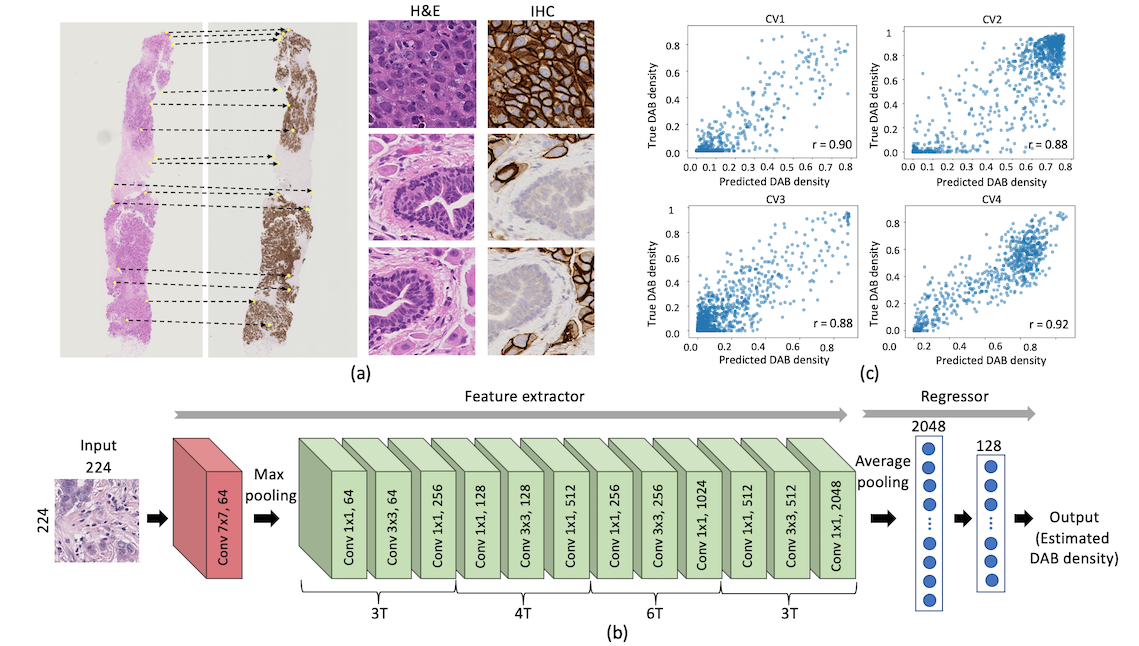}
\end{center}
\caption{DAB density estimation using a small subset of the HER2C dataset \citep{qaiser2018her}: (a) Image registration between H\&E and corresponding IHC images using control points (yellow points \textit{left}) on both WSIs; \textit{right}: three examples of H\&E patches and corresponding registered IHC patches; (b) Convolutional neural network architecture for regressing DAB density from H\&E images. 'T' represents the number of resnet blocks used; (c) Scatter plots between model prediction and true DAB density using 4-fold cross validation (averaged Pearson correlation coefficient 0.90 with $p$-value $<$ 0.0001).} \label{registration_HE_IHC}
\end{figure*}

\begin{figure*}[h]
\begin{center}
\includegraphics[width=1.0\textwidth]{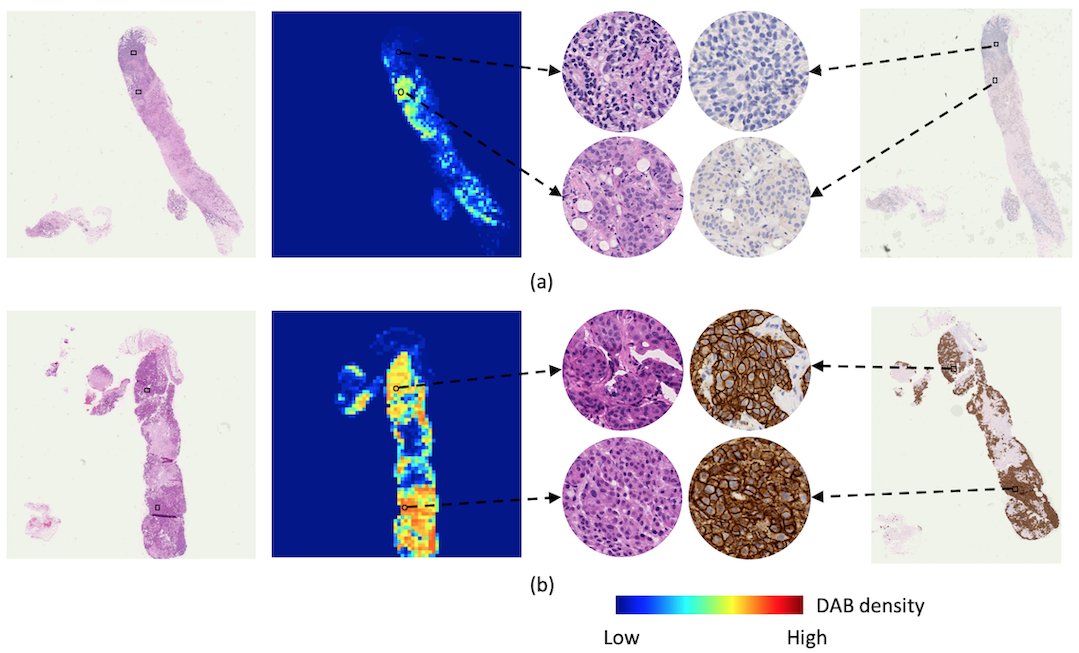}
\end{center}
\caption{Visualisation of the estimated DAB density on test WSIs from the HER2C dataset \citep{qaiser2018her}: (a) HER2-negative case; (b) HER2-positive case. From left to right column: raw H\&E stained WSI; estimated DAB density using the trained regression model; zoomed-in version of two local regions; corresponding IHC stained WSI as the reference. } \label{overlay_regression}
\end{figure*}

\subsection{Feature extraction}
Given a stain normalised patch $x_j$, a variety of representative features $\boldsymbol{h}_j$ can be extracted. The objective of this feature extraction step is to obtain features that are associated with tissue characteristics in the patch and the target variable of interest as well. These features include nuclear composition features (e.g., counts of different types of nuclei in the patch), morphological features, receptor expression features (\hyperref[sec:DDE]{section 2.2.3}), deep features (or neural feature embdeddings from a pre-trained neural network) and a combination of these. We explore the association between different kinds of features and the HER2 status. It is important to note that the proposed framework is generic and not restricted to any particular type of features and the graph neural network (discussed later) can be used to accumulate a variety of node level features.

\subsubsection{Nuclear composition features (NCF)}
HER2 status has been shown to be associated with the presence of different types of nuclei in BCa tissues \citep{lu2020capturing}. Here, we extract nuclear composition features which cover the counts of nuclei of different types of cells in a patch. Specifically, for a given patch $x_j$, we use HoVer-Net \citep{graham2019hover} trained on the BCa PanNuke dataset \citep{gamper2019pannuke} to localise nuclei and predict their types. HoVer-Net is a convolutional neural network for simultaneous nuclear segmentation and classification. This network leverages instance-rich information encoded within vertical and horizontal distance maps of nuclear pixels to their centres of mass and achieves accurate segmentation even in areas with overlapping instances. Five categories of nuclei are predicted: nuclei of neoplastic, non-neoplastic epithelial, inflammatory, connective tissue and necrotic cells. For each image patch, HoVer-Net generates a set of nuclear centroids together with their cell type and the corresponding nuclear segmentation mask. Nuclear composition features in a patch can then be collected by counting the number of the five types of nuclei in it.

\subsubsection{Nuclear morphological features (NMF)}
Assuming that the morphology of different types of nuclei is associated with HER2 status, we extract 15 nuclear morphological features such as nuclear size, eccentricity, orientation, and length of the major axis (see Table~\ref{feature_15_explain}) using the output binary mask of each nucleus. Therefore, each detected nucleus is represented by a 15-dimensional feature vector which contains 15 different morphological properties. 
We use the mean and standard deviation of the 15 feature values resulting in a 30-dimensional feature vector for each patch. 

\subsubsection{Neural embeddings (NE)}
One of the strengths of deep neural networks is their ability to learn high-level features based on colour, frequency domain, edge detectors, texture and so on from image pixels. An image patch is fed into the network and transformed several times through convolutional layers in the network. During these transformations, the network is able to learn new and increasingly complex features of the input image. In order to extract a strong and representative set of features, we extract features from the last convolutional layer of ResNet50 \citep{he2016deep} due to its excellent performance in recent computer vision tasks. The model was trained on the ImageNet dataset \citep{deng2009imagenet} which is a large visual database designed for visual object recognition research. 

\subsubsection{DAB density estimates (DDE)}\label{sec:DDE}

Where IHC images are available, areas of membranous DAB staining reveal the level of HER2 protein expression. Based on this, we utilised paired H$\&$E and IHC images in the HER2C dataset to develop a deep neural network predictor to estimate the level of HER2 expression (DAB density estimates) in a given H$\&$E image region which is then used as a node-level feature in the graph neural network. First, we performed affine registration on 5 H\&E and IHC paired images from the HER2C dataset \citep{qaiser2018her} by taking the H\&E WSIs as the reference image and extracting several control points pairs at 40X resolution from each image pair. Fig.~\ref{registration_HE_IHC} (a) shows an example H$\&$E and IHC pair and the corresponding control points (yellow dots). The calculated affine transform matrix, consisting of rotation, scaling and translation, is applied on all H\&E image patches to get the corresponding IHC images. Fig.~\ref{registration_HE_IHC} (a) gives three examples of the H\&E and registered IHC images which shows high registration accuracy even at the highest resolution (40$\times$). Second, in the registered IHC images, we convert their RGB colour space to Haematoxylin-Eosin-DAB (HED) colour space and calculate the percentage of DAB staining from the DAB channel. 

In total, we collect more than 6,000 H\&E patches (size 224$\times$224 pixels) and their corresponding DAB density values. Architecture of the proposed regression model is shown in Fig.~\ref{registration_HE_IHC}(b). The feature extraction component of the network is inspired by ResNet50. Compared to the standard ResNet50 implementation, we add two fully connected layers after the feature extraction component with 2048 and 128 neurons, respectively. In order to evaluate the performance of our regression model, we do a leave-one-WSI-out cross validation using the collected dataset and calculate the Pearson correlation coefficient (PCC) to measure the linear correlation between the ground truth and model prediction (Fig.~\ref{registration_HE_IHC}(c)). Strongly positive correlation can be observed in all the 4 folds, achieving the averaged PCC 0.90 ($p$ $<$ 0.0001). Fig.~\ref{overlay_regression} shows the visualisation of the estimated DAB density on a HER2-negative (first row) and a HER2-positive (second row) case, respectively. The WSIs showed in Fig.~\ref{overlay_regression} are unseen by the trained model. It can be observed from the HER2-negative case that the majority of the tissue region has low estimated DAB density, revealing the lack of HER2 protein expression. Compared to the negative case, the HER2-positive case has larger areas with high estimated DAB density, as observed from the orange and red areas in the heatmap. The highlighted activation areas (zoomed-in) in the generated heatmap are consistent with the DAB density in the corresponding IHC images. This supports the idea of using DAB density as a potential feature for HER2 status prediction from H\&E WSIs in computational pathology.

The above features are used for graph construction using \textbf{Algorithm 1} as described below.

\hypertarget{Algorithm 1}{}
\begin{table}[!t]
\centering
\begin{tabular}{p{7.5cm}}
\hline
\textbf{Algorithm 1}: WSI Graph Construction.\\
\hline
\vspace{-5pt}
\textbf{INPUT}: A set of $n$ patches in a given WSI - each represented by its spatial coordinates $\boldsymbol{g}_j$ and feature vector $\boldsymbol{h}_j$. \\
\textbf{OUTPUT}: Graph representation $\boldsymbol{G}=(\boldsymbol{V},\boldsymbol{E})$ of the WSI. \\
\textbf{PARAMETERS}: \\
\quad $s_{min}=0.8$, $d_{max}=4000$ pixels at 0.25mpp \\
\textbf{STEPS}:   \\
1: Perform agglomerative clustering using the similarity metric $k(\cdot , \cdot)$ with average linkage. Cluster agglomeration takes place up to a minimum similarity threshold of $s_{min}$. For a given WSI $X$, this results in a cluster set $C(X)$ with each patch $x_j\equiv \left ( \boldsymbol{g}_j, \boldsymbol{h}_j \right ) \in X$ assigned to exactly one cluster $c \in C(X)$. \\
2: For each cluster set $c$, compute the geometric centre of all its constituent patches $\boldsymbol{p}_c=\frac{1}{|c|} \sum_{\boldsymbol{(g_j,h_j)}\in c} \boldsymbol{g}_j$. \\
3: For each cluster set $c$, compute the aggregated feature vector of all its constituent patches $\boldsymbol{u}_c=\frac{1}{|c|} \sum_{\boldsymbol{(g_j,h_j)}\in c} \boldsymbol{h}_j$. \\
4: Construct a vertex set $V$ for the given WSI $X$ with each cluster $c\in C(X)$ as a node represented by $\boldsymbol{v}_c \equiv (\boldsymbol{p}_c, \boldsymbol{u}_c)$. \\
5: Use Delauney triangulation to construct the edge set $E$ based on the geometric coordinates of cluster centres with a maximum distance connectivity threshold of $d_{max}$ pixels. \\
\hline
\end{tabular} 
\label{algorithm1}
\end{table}

\subsection{Adaptive Spatial Agglomerative Clustering}
As the number of patches in a WSI can be quite large, we group spatially neighbouring regions with high degree of similarity in the feature space in order to reduce the computational cost of downstream analysis. This is achieved using adaptive spatial agglomerative clustering which relies on a patch-level similarity kernel (see \textbf{Algorithm 1}). We use a feature space Gaussian kernel $k_h \left (\boldsymbol{h}_{a},\boldsymbol{h}_{b} \right )={\rm{exp}}\left ( {-{\lambda}_h\|\boldsymbol{h}_{a}-\boldsymbol{h}_{b}\|} \right )$ and a geometric Gaussian kernel $k_g \left (\boldsymbol{g}_{a},\boldsymbol{g}_{b} \right )={\rm{exp}}\left ( {-{\lambda}_g\|\boldsymbol{g}_{a}-\boldsymbol{g}_{b}\|} \right )$ to determine the degree of similarity between feature space representations and geometric coordinates of two patches in a given WSI (indexed by $a$ and $b$), where ${\lambda}_h$ and ${\lambda}_g$ control the degree of similarity in feature space and distance in the geometric coordinates, respectively. We use a joint product kernel $k\left(x_a, x_b\right)=k_h \left (\boldsymbol{h}_{a},\boldsymbol{h}_{b} \right )k_g \left (\boldsymbol{g}_{a},\boldsymbol{g}_{b} \right )$ to model the overall similarity between two patches in a WSI. Note that the kernel product implies a tensor product of underlying features. The pairwise similarity matrix of all patches in a given WSI is then used as a similarity metric in Agglomerative clustering \citep{mullner2011modern} with average linkage. This is done such that cluster agglomeration takes place up to a minimum similarity threshold of $s_{min}$. For a given WSI $X$, this results in a cluster set $C(X)$ with each patch $x_j\equiv \left ( \boldsymbol{g}_j, \boldsymbol{h}_j \right ) \in X$ assigned to exactly one cluster $c \in C(X)$. It is important to note that each cluster contains a number of patches and the number of clusters across all the WSIs can be different depending upon the sizes and morphological complexity of the WSIs.

\begin{figure*}[htp]
\begin{center}
\includegraphics[width=1\textwidth]{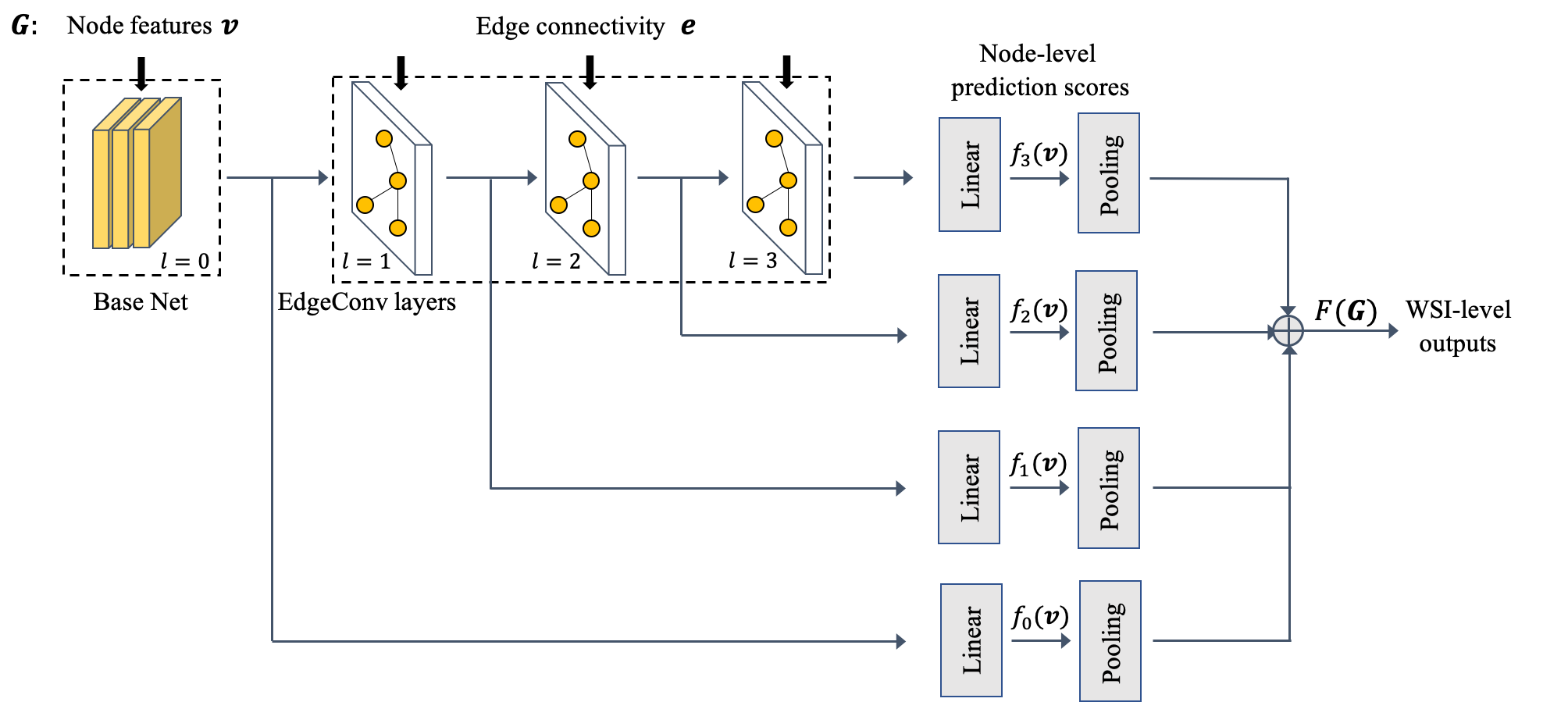}
\end{center}
\caption{Architecture of the proposed SlideGraph$^{+}$ model for graph based WSI classification. The Base Net block is composed of a convolutional layer, Batch Normalisation and Rectified Linear Unit activation (RELU) layers. Graph neural network layer is structured with edge convolution (EdgeConv) layers whose mathematical expression is shown in (\ref{eq:RMPB}). Pooling, Linear refer to the pooling and linear layers. } \label{GCN_architecture}
\end{figure*}

\subsection{Graph construction}
We construct a WSI-level graph representation $G(X)\equiv(V,E)$ of a given WSI $X$ based on its clusters \citep{sharma2015review,lu2019new,lu2019graph}. The graph representation of a WSI $X$ consists of a vertex set $V$ with each cluster $c\in C(X)$ as a node represented by $\boldsymbol{v}_c \equiv (\boldsymbol{p}_c, \boldsymbol{u}_c)$ based on the geometric centre $\boldsymbol{p}_c=\frac{1}{|c|} \sum_{\boldsymbol{(g_j,h_j)}\in c} \boldsymbol{g}_j$ and aggregated feature vector of all its constituent patches $\boldsymbol{u}_c=\frac{1}{|c|} \sum_{\boldsymbol{(g_j,h_j)}\in c} \boldsymbol{h}_j$. The edge set $E \subseteq V \times V$ represents a finite set of edges between nodes. In order to capture communication patterns between components of the tissue, the edge set is constructed by using Delauney triangulation based on the geometric coordinates of cluster centres with a maximum distance connectivity threshold of $d_{max}$ pixels \citep{chew1989constrained}. This results in a planar graph, i.e., no two edges in the graph intersect each other.

\subsection{GNN architecture}
The proposed architecture uses a Graph Neural Network (GNN) to generate both regional (node-level) and WSI-level predictions based on the above WSI-level graph representation. GNNs have the advantage of being inherently invariant to rotation and translation of graph nodes \citep{chami2020machine} and can learn progressively abstract node-level feature embdeddings across their layers through message passing or information sharing between neighbouring graph nodes. The architecture of the proposed graph neural network is shown in Fig.~\ref{GCN_architecture}. It consists of multiple edge-convolution (EdgeConv) layers with each layer using a multi-layer perceptron (MLP) to produce a feature embedding of a node in the graph based on the feature embeddings of the node itself and its neighbouring nodes generated by previous layers \citep{wang2019dynamic}. The first layer uses the original node-level features discussed above. The MLP in each layer uses the embedding of a node and the difference of its node embedding from its neighbours. Consequently, each GNN layer accumulates information from progressively higher order neighbours of each node. Mathematically, the output feature representation of an EdgeConv layer $l=1 \ldots L$ in the GNN with $L$ layers for a given node at index $k$  in the input graph can be written as:

\begin{equation}\label{eq:RMPB}
\boldsymbol{u}^{\left ( l \right )}_k = \sum_{j \in N_k}
        H^{\left ( l \right )}(\boldsymbol{u}^{\left ( l-1 \right )}_k \, , \,
        \boldsymbol{u}^{\left ( l-1 \right )}_j - \boldsymbol{u}^{\left ( l-1 \right )}_k;{\theta_l}).
\end{equation}

In the above equation, $\boldsymbol{u}^{\left ( 0 \right )}_k=\boldsymbol{u}_k$, $N_k$ denotes the neighbourhood of node $k$ and $H^{\left ( l \right )}$ represents the multi-layer perceptron (MLP) with trainable weights $\theta_l$. As shown in Fig.~\ref{GCN_architecture}, the feature embedding $\boldsymbol{u}^{\left ( l \right )}_k$ of a node $\boldsymbol{v}_k \equiv (\boldsymbol{p}_k, \boldsymbol{u}_k)\in V$ from a GNN layer is passed to a corresponding linear layer to generate node level predictions $f_l(\boldsymbol{v}_k)=\boldsymbol{w}_l^T \boldsymbol{u}^{\left ( l \right )}_k $. These node level prediction scores are then pooled to generate layer-wise WSI-level predictions, i.e., $F_l{(\boldsymbol{G})}=\sum_{\forall v\in \boldsymbol{V}}f_l(v)$. These scores are then summed to produce the overall WSI-level prediction score, i.e.,  $F{(\boldsymbol{G};\theta)}=\sum_{l=0}^L F_l{(\boldsymbol{G})}$ with all trainable parameters $\theta$. It is important to note that SlideGraph$^{+}$ uses node-level aggregation as opposed to SlideGraph (\cite{lu2020capturing}) which only has WSI-level output.

In this work, we use three EdgeConv layers ($L=3$) which have 16, 16 and 8 neurons respectively in their MLP with a linear layer, followed by a batch normalisation (BN) and a Rectified Linear Unit activation (RELU) layers. The code is available at \url{https://github.com/wenqi006/SlideGraph} for further studies.

\begin{table*}[htp]
\caption{A comparison of SlideGraph$^{+}$ and the state-of-the-art methods on datasets from multiple centres. For all the results shown, the training is always done using the TCGA-BRCA dataset.}\label{Results_comparision}
\resizebox{\textwidth}{!}{%
\begin{tabular}{llll}
\hline
Method         & Feature (dimension)                  & Test set                                                         & AUROC (mean$\pm$std)           \\\hline
\citep{kather2019pan}     &   Shufflenet                                      & TCGA-BRCA                                                             &  0.62                             \\
\citep{rawat2020deep}    &  Fingerprints (512)                                   & TCGA-BRCA                                                             &   0.71$\pm$0.04                               \\\hline
Linear regression & Maximum DAB density estimates (1)               & TCGA-BRCA                                                             & 0.593                   \\
               & Majority DAB density estimates (1)                   & TCGA-BRCA                                                             & 0.573                            \\
               & Average DAB density estimates (1)                    & TCGA-BRCA                                                             & 0.598                             \\\hline
SlideGraph$^{+}$      & Nuclear composition (5)                           & TCGA-BRCA                                                             & 0.71$\pm$0.02 \\
(Cross validation on               & Cellular morphology (30)                         & TCGA-BRCA                                                             & 0.72$\pm$0.05 \\
 TCGA-BRCA dataset)              & Resnet50 (2048)                            & TCGA-BRCA                                                             & 0.72$\pm$0.07  \\
               & DAB density estimates (4)                           & TCGA-BRCA                                                             & \textbf{0.75$\pm$0.02}                         \\
               & Nuclear composition + Cellular morphology (35)            & TCGA-BRCA                                                             & 0.75$\pm$0.04  \\
               & Nuclear composition + Cellular morphology + DAB (39)         & TCGA-BRCA                                                             & 0.75$\pm$0.08  \\\cline{1-4} 
SlideGraph$^{+}$                & DAB density estimates (4)                             & \begin{tabular}[c]{@{}l@{}}HER2C (0 / 3+)\end{tabular}         & \textbf{0.78$\pm$0.03}                                  \\
(Independent validation      &        & \begin{tabular}[c]{@{}l@{}}HER2C (0, 1+ / 3+)\end{tabular}     & 0.75$\pm$0.02                               \\
 on HER2C and               &                & \begin{tabular}[c]{@{}l@{}}Nott-HER2 (0 / 3+)\end{tabular}     & \textbf{0.80$\pm$0.02}                                  \\
Nott-HER2 datasets)               &                  & \begin{tabular}[c]{@{}l@{}}Nott-HER2 (0, 1+ / 3+)\end{tabular} & 0.79$\pm$0.01                               \\
               &                   & \begin{tabular}[c]{@{}l@{}}Nott-HER2 (- / +)\end{tabular}      & 0.71$\pm$0.01              \\ \hline  

\end{tabular}}
\end{table*}

\subsection{Loss function and hyperparameter settings}
The proposed SlideGraph$^{+}$ is implemented using the PyTorch Geometric (PyG) library \citep{fey2019fast} \citep{paszke2017automatic}. During the training, WSI-level prediction score is compared to the WSI-level target and weights in the EdgeConv layers and the base neural network block are updated through backpropagation. In each training batch, a set of positive cases $B^+$ and a set of negative cases $B^-$ are chosen in a stratified manner. The loss function is designed as a pairwise ranking based hinge-loss function with the mathematical formulation as follows:
\begin{equation}
\mathcal{L}(B^+,B^-;\theta) = \sum_{{i \in B^+}}\sum_{{j \in B^-}}max{\left (0, 1 - (F(G_i; \theta) - F(G_j; \theta)) \right )}.
\end{equation}

The minimisation of the loss function is implemented by using adaptive momentum-based optimisation \citep{kingma2014adam} with the learning rate 0.001 and a weight decay 0.0001. After training, the performance of the predictor is evaluated over test datasets. We use Area under Receiver Operator Characteristic (AUROC) curve and Precision-Recall (AUPR) curve to evaluate the predictive performance over test sets \citep{davis2006relationship}.

\section{Results and discussion}

\subsection{HER2 status prediction}
The current published state-of-the-art method by Rawat et al. \citep{rawat2020deep} gives AUROC values of 0.71 under five-fold cross-validation. In line with previous methods, we report AUROC for comparison. Table~\ref{Results_comparision} shows the AUROC values by the proposed SlideGraph$^{+}$ and existing state-of-the-art methods. Using the same cross-validation strategy \citep{rawat2020deep}, our proposed SlideGraph$^{+}$ model with DAB density estimates achieves the best AUROC with 0.75$\pm$0.02. It can be observed that all feature compositions under the SlideGraph$^{+}$ framework achieve higher AUROC values than the state-of-the-art methods. Among them, feature combinations `Nuclear composition + Nuclear morphology + DAB density estimates', `Nuclear composition + Nuclear morphology' and `DAB density estimates' exceed the state-of-the-art by a significant margin, obtaining the maximum AUROC value of 0.75. 
\begin{figure}[h!]
\begin{center}
\includegraphics[width=0.45\textwidth]{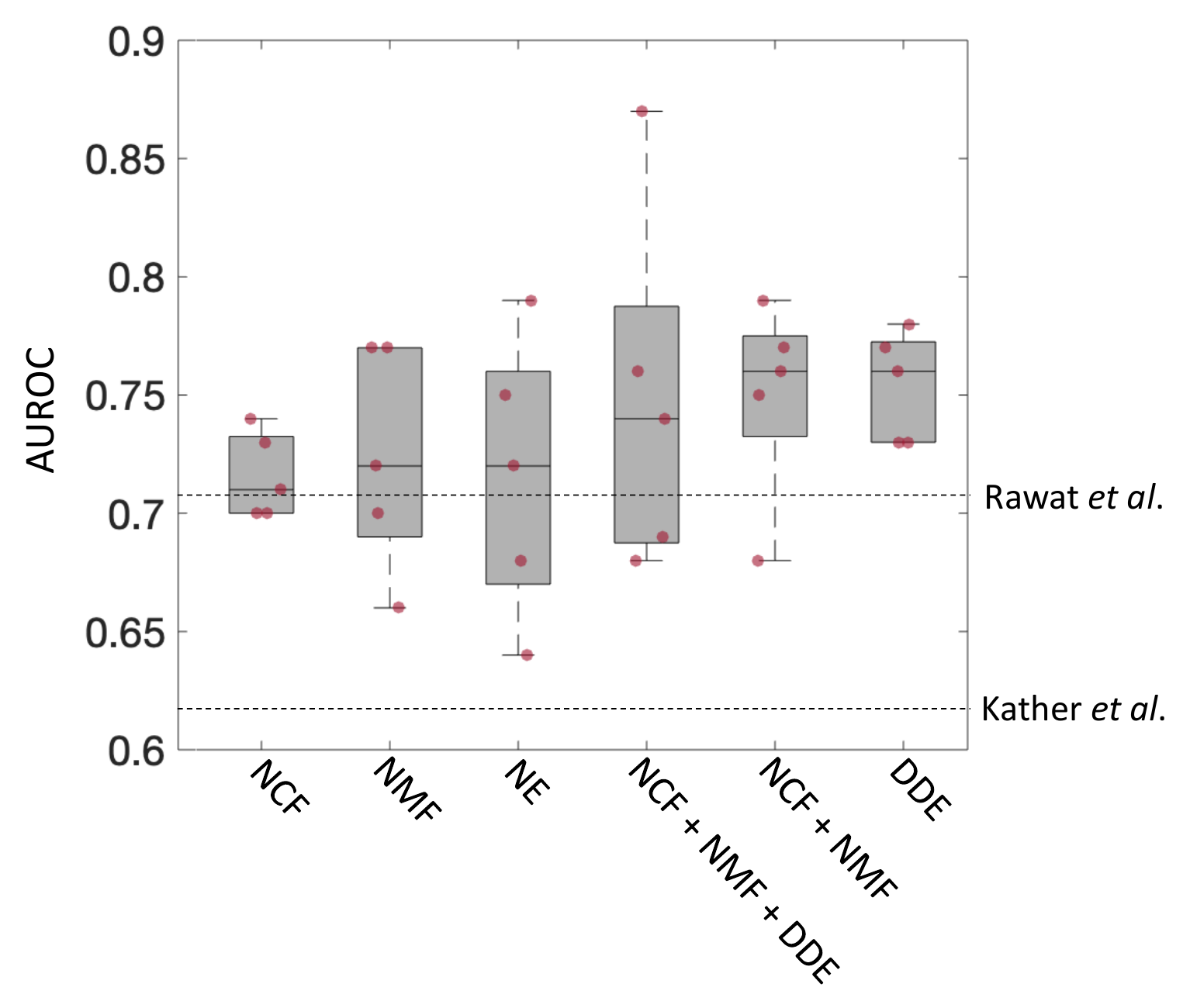}
\end{center}
\caption{AUROC values using different feature compositions during the five-fold cross validation on TCGA-BRCA. Dashed lines show the state-of-the-art AUROC values. NCF: Nuclear composition features; NMF: Nuclear morphological features; NE: Neural embeddings; DDE: DAB density estimates.} \label{boxplots_differentfeatures}
\end{figure}
The SlideGraph$^{+}$ model with DAB density estimates achieves the smallest standard deviation with 0.02 in AUROC, proving its stability in HER2 status prediction. In addition, the estimated DAB density feature only has 4 dimensions, leading to the fewest number of training parameters and highest computational efficiency.
\begin{figure*}[h!]
\begin{center}
\includegraphics[width=1\textwidth]{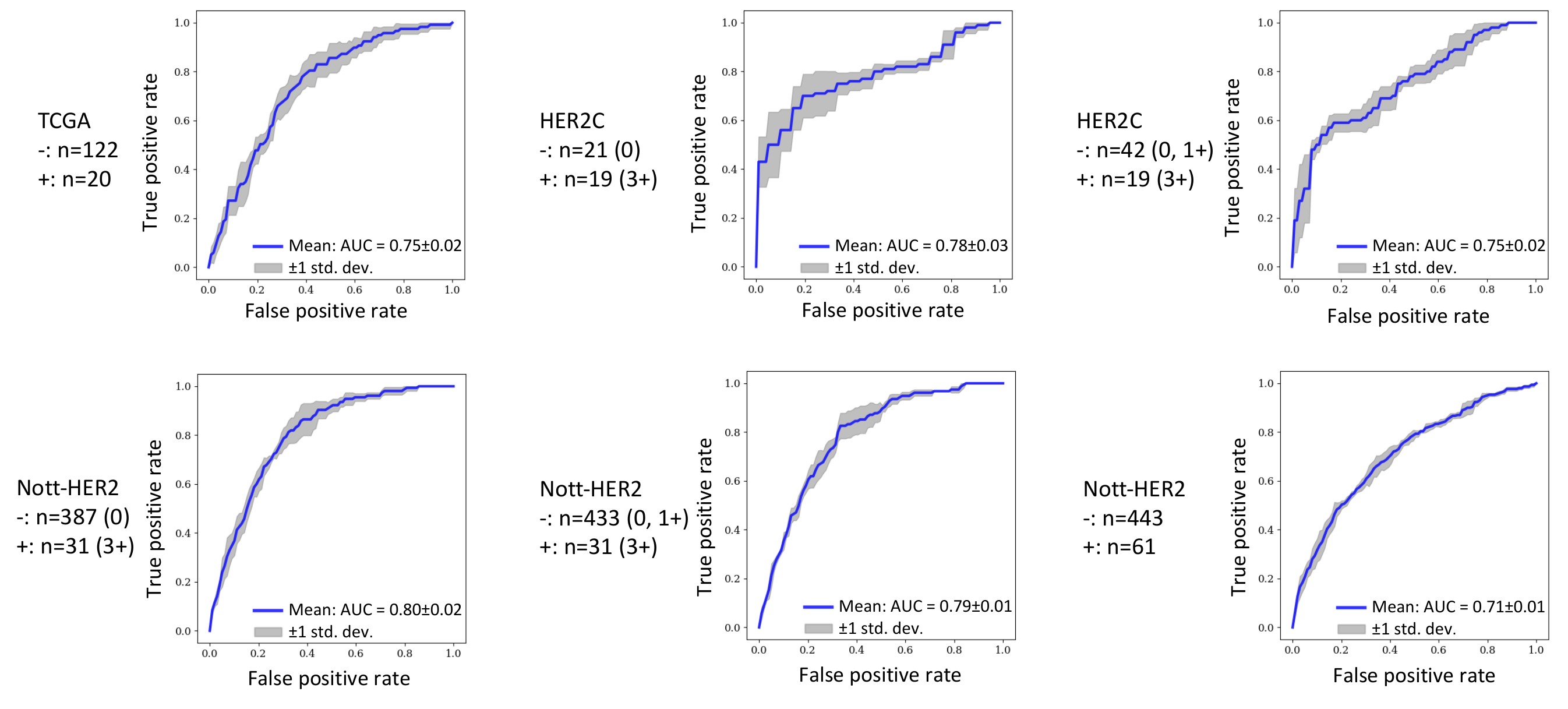}
\end{center}
\caption{ROC curves when testing the trained SlideGraph$^{+}$ classification model on the TCGA test dataset and on the other two independent test datasets (HER2C and Nott-HER2)} \label{testresults}
\end{figure*}
Distribution of the AUROC values achieved using different feature compositions are shown in Fig.~\ref{boxplots_differentfeatures}. 

In order to compare the performance of the GNN in comparison to a naive aggregation of DAB density prediction scores, we average the estimated DAB densities from the four trained regression models on each patch and use three aggregating strategies (average pooling, max pooling and majority voting) to generate the overall WSI-level DAB density. Here we confine the average pooling and majority voting strategies on patches whose estimated DAB density is higher than 0.1. We calculate three type of DAB features -- namely maximum DAB density estimates, majority DAB density estimates and average DAB density estimates respectively -- on each WSI. As can be seen from Table~\ref{Results_comparision}, among all the three DAB density estimates features, the average DAB density estimates obtains the highest AUROC 0.598. The above results demonstrate the superiority of our proposed SlideGraph$^{+}$ architecture. Combining DAB density estimates with the help of a graph gives much higher AUROC and better HER2 prediction performance than the DAB density estimates on its own. 

\subsection{Independent validation}
We then test our trained model on two independent test datasets: HER2C and Nott-HER2. Here, we utilise the model trained using DAB density estimates due to its superior performance and simplicity. As can be seen from Table~\ref{Results_comparision}, on HER2C dataset, the model achieves mean AUROC of 0.78 when differentiating negatives cases (status 0) and positive cases (3+). When we add 1+ cases to the negative group, our trained model achieves mean 0.75 AUROC value. For the Nott-HER2 dataset, our trained model achieves mean AUROC of 0.80 (0/3+), 0.79 (0, 1+/3+) and 0.71 (-/+) respectively. Corresponding ROC curves can be seen in Fig.~\ref{testresults}. The independent validation on multi-centre datasets demonstrates the generalisation ability of the proposed SlideGraph$^{+}$ model. 

\subsection{Performance comparison using AUPR}

\begin{figure*}[h!]
\begin{center}
\includegraphics[width=0.8\textwidth]{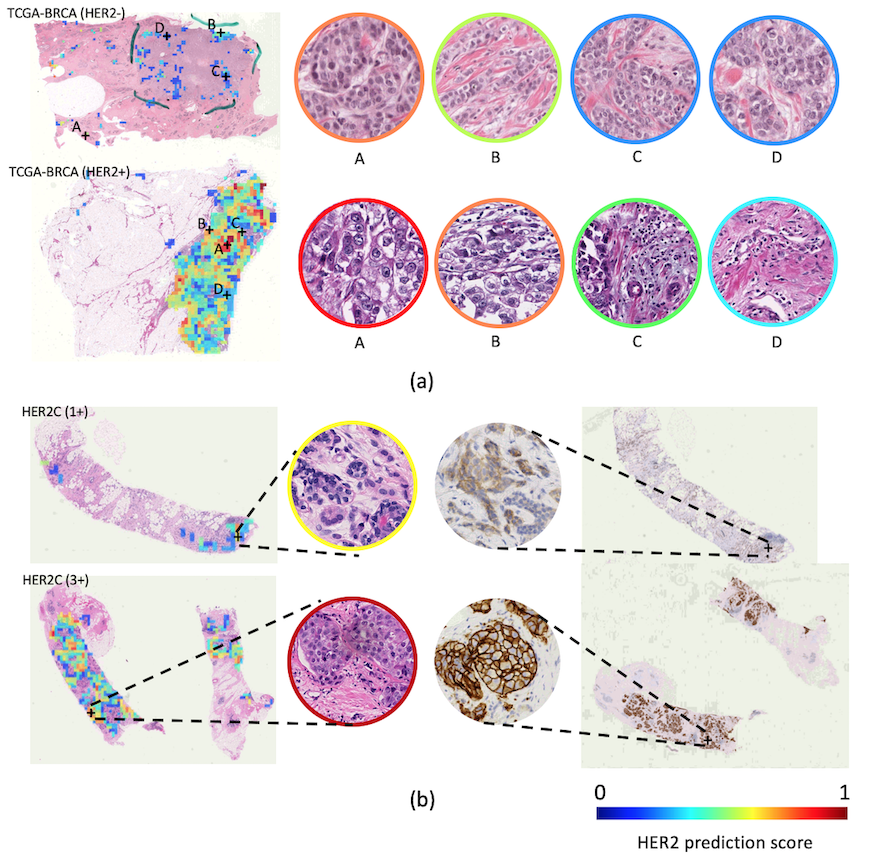}
\end{center}
\caption{Example heatmaps of node-level prediction scores: (a) cases from TCGA-BRCA; (b) cases from HER2C. Top row: HER2 negative; Bottom row: HER2 positive. Boundary colour of each zoomed-in region represents its contribution to HER2 positivity (prediction score). A-D in (a) denotes the positions of the zoomed-in regions. IHC images in (b) illustrate that the H\&E regions with high HER2 prediction scores are consistent with the strongly stained DAB areas in the corresponding IHC images.} \label{node_visulisation}
\end{figure*}

\begin{figure*}[htp]
\includegraphics[width=1\textwidth]{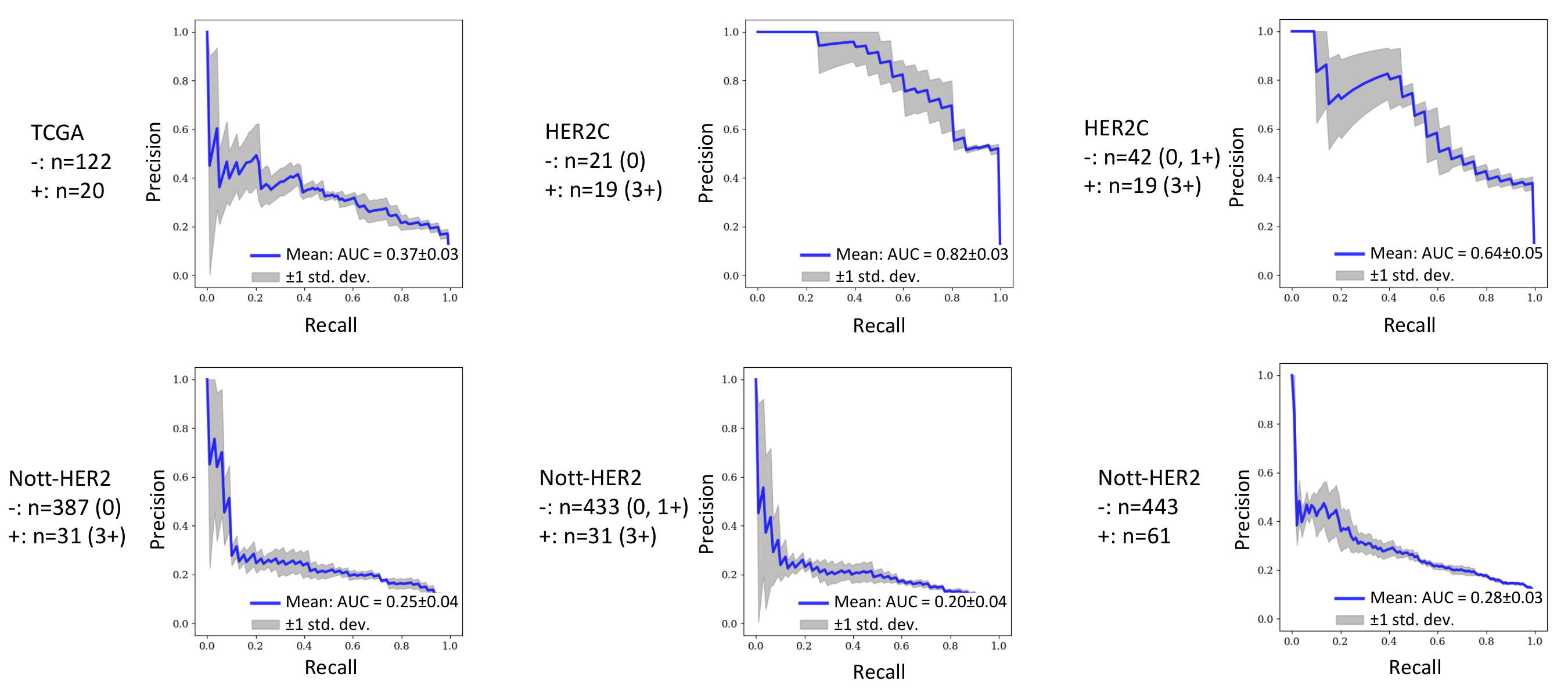}
\caption{PR curves when testing the trained SlideGraph$^{+}$ classification model on the TCGA-BRCA test dataset and on the other two independent test datasets (HER2C and Nott-HER2)} \label{testresults_aupr}
\end{figure*}

Despite significant class imbalance, the previously published works \citep{kather2019pan, rawat2020deep} did not report AUPR results. In this paper, in addition to AUROC, we also record the AUPR results in each experimental setting and show the values in Table~\ref{Results_comparision_AUPR} and Fig.~\ref{testresults_aupr}. When doing the five-fold cross-validation on the TCGA-BRCA dataset, the SlideGraph$^{+}$ model with DAB density estimates achieves the best AUPR 0.37$\pm$0.03. Sole DAB density estimate feature achieves the maximum AUPR value of 0.20, showing the significant impact of our proposed SlideGraph$^{+}$ architecture. We also test our trained model on the other two independent test datasets. On the HER2C dataset, the model achieves mean AUPR 0.82 when differentiating definitive negative cases (status 0) and positive cases (3+). When we add 1+ cases to the negative group, our trained model achieves mean AUPR value of 0.64. For the Nott-HER2 dataset, our model achieves mean AUPR 0.25 (0/3+), 0.20 (0, 1+/3+) and 0.28 (-/+) respectively.

\begin{figure*}[h!]
\begin{center}
\includegraphics[width=1.0\textwidth]{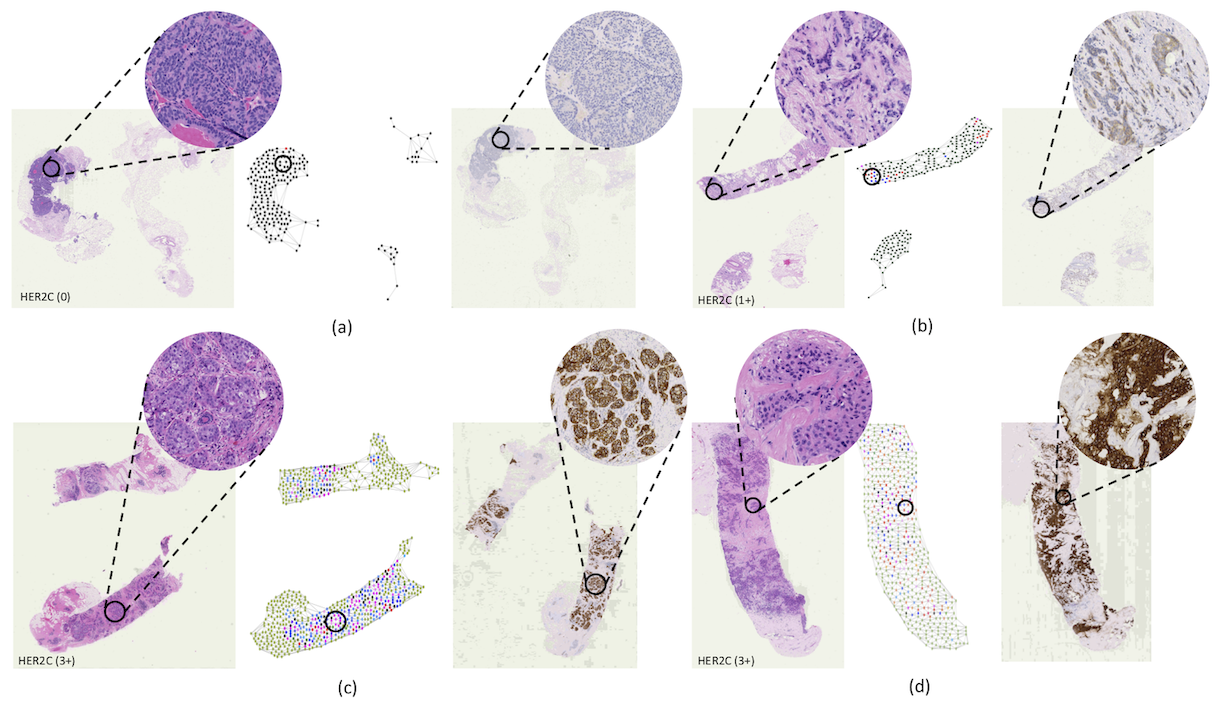}
\end{center}
\caption{Visualisation of node-level prediction on independent HER2C test dataset. Top row: HER2-negative; Bottom row: HER2-positive. } \label{more_node_visulisation}
\end{figure*}
It may be worth noting that AUPR in the independent test sets are not comparable because the number of cases in both classes and the ratio between them varies among different settings.

\subsection{Visualisation of HER2 predictions}
In order to understand the ability of WSI-level graphs to capture tissue architecture and their predictive power for WSI-level prediction of receptor status, let us examine the node-level prediction performance on several cases from TCGA-BRCA and HER2C datasets. Fig.~\ref{node_visulisation}(a) shows the overlay of heatmaps and four zoomed-in regions which have different levels of HER2 prediction score. It can be observed that only a few areas in the negative sample contribute to the HER2 positivity while majority of the tissue regions in the positive case have high HER2 prediction scores. Same can be observed on sample images from the HER2C dataset in Fig.~\ref{node_visulisation}(b). It should also be noted that regions with high HER2 prediction scores are consistent with high DAB intensity areas in the corresponding IHC images.

We then convert the node-level prediction score into a false color representation of each node. This results in a WSI-level graph visualisation in which the colour of each node is based on its node-level prediction score. Fig.~\ref{more_node_visulisation} shows the results of this visualisation for two HER2-negative (top row) and two HER2-positive (bottom row) WSIs. One can observe clear differences in the graphs of the two classes: note the prevalence of red and blue areas in HER2-positive WSIs and dark green areas in HER2-negative WSIs. This supports the overall idea of using WSI-level graphs proposed in this work and the utility of incorporating global context for machine learning problems in computational pathology.

\begin{figure*}[h!]
\centering
\includegraphics[width=0.65\textwidth]{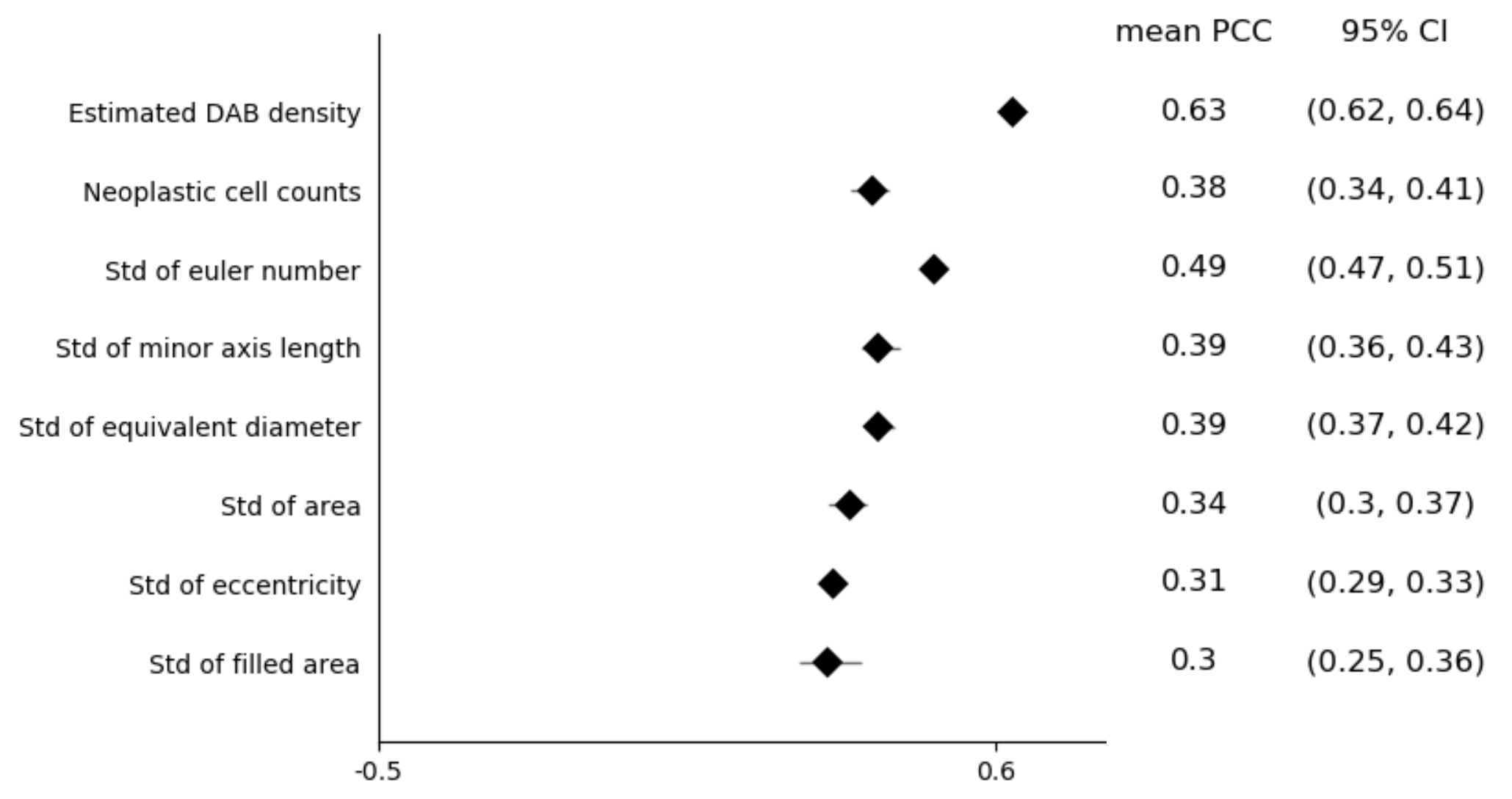}\\
\includegraphics[width=1.0\textwidth]{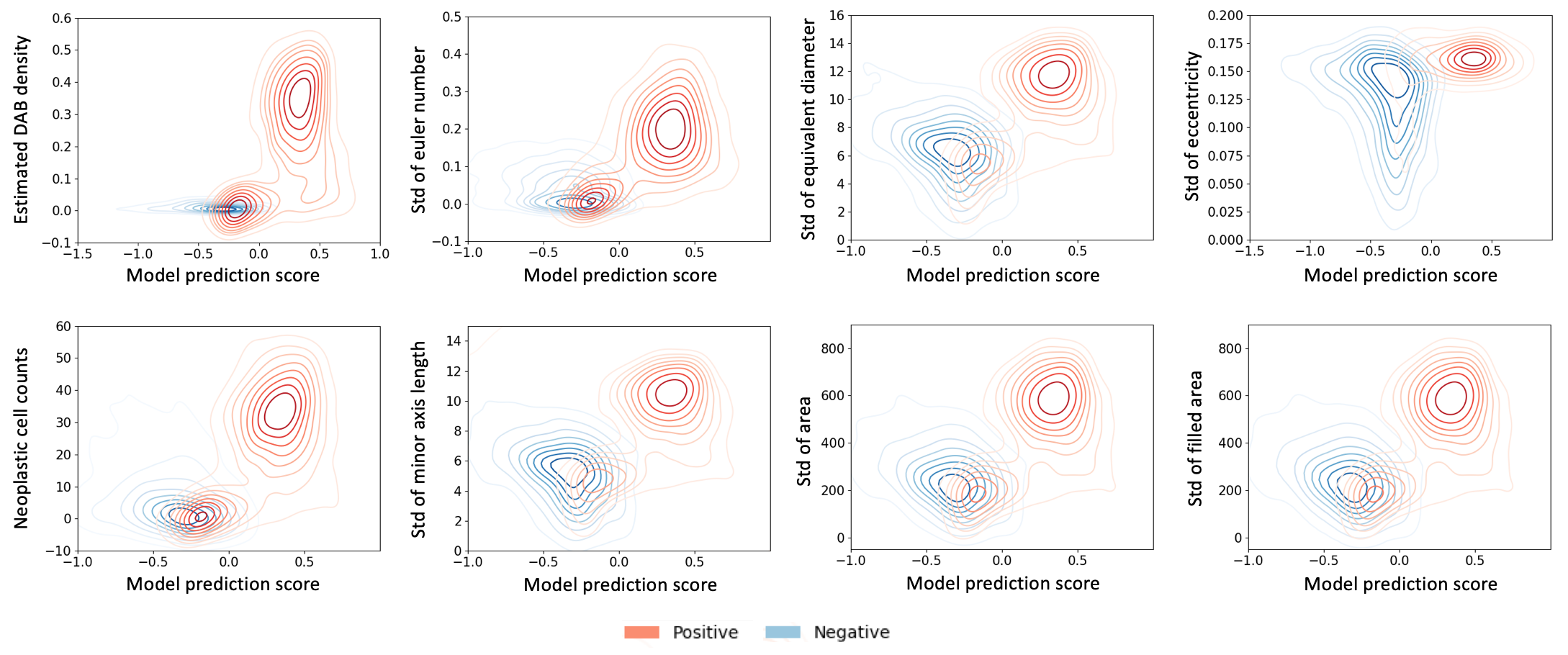}
\caption{Top: Correlation between nuclear pleomorphism and model prediction scores for ten patients from TCGA-BRCA cohort. Each row gives the mean and confidence interval (CI) of PCC between the graph node-level prediction score and a specific feature. For all features, $p<$ 0.0001. Bottom: Density plots of nuclear pleomorphism related features and model prediction scores between positive cases (red) and negative cases (blue).} \label{pleomorphism2_combine}
\end{figure*}

\subsection{Correlation between nuclear pleomorphism and model prediction score}
We conduct further analysis on nuclei pleomorphism related features that contribute to the HER2 prediction. We include five HER2-positive and five HER2 negative-cases in this experiment and calculate the Pearson correlation coefficient (PCC) between the node-level prediction score and cell nuclei pleomorphism related features. In Fig.~\ref{pleomorphism2_combine} (top), we show the mean, confidence interval (CI) of PCC and limit our discussion to features whose mean PCC is above 0.3. We can see that the estimated DAB density feature gives the highest PCC value of 0.63 (95\% CI 0.62, 0.64). Among the nuclear composition features, neoplastic cell counts contribute more to HER2 positivity prediction, with mean PCC 0.38 and 95\% CI (0.34, 0.41). This is plausible because regions with HER2 over-expression normally have larger number of neoplastic cells.

Among all the nuclear pleomorphism related features, standard deviation of the Euler number gives the strongest positive correlation value of 0.49 (95\% CI 0.47-0.51, $p<$0.0001). Mathematically, in a 2D nuclear mask, the Euler number is the number of objects minus the number of holes. Hence, the standard deviation of the Euler number may capture the diversity of nuclear morphology and chromatin texture. Higher values represent major morphological differences. Standard deviation values of minor axis length, equivalent diameter, area, eccentricity and filled area are another five nuclear pleomorphism related features which give mean PCC above 0.3 with $p <$ 0.0001. These five features are associated with the significance of variation in shape and sizes of the cells. Density plots of features and model prediction scores between positive cases (red) and negative cases (blue) are shown in Fig.~\ref{pleomorphism2_combine} (bottom). Clear separation can be observed between the positive and negative groups. The observations here point to the association of nuclear pleomorphism with HER2 positivity and cancer progression.

\subsection{Comparison of computational efficiency}
We have also compared the computational efficiency of patch-based and the proposed SlideGraph$^{+}$ model using a single Nvidia Titan RTX GPU. 
Once the patches and graphs are ready from the WSI, the average single-fold training time for the baseline model \citep{kather2019deep} is 5.3 hours and the testing time for a WSI is 1.2 seconds from patches to the final prediction whereas SlideGraph$^{+}$ training for a single fold takes 2 minutes on average and 0.4 milliseconds to get the label prediction from a single graph. 




\section{Conclusions}
In this paper, we proposed SlideGraph$^{+}$, a generic method that couples WSI-level graph representation with a graph convolutional network for capturing the global context of a WSI and showed its effectiveness for prediction of HER2 status directly from WSIs of H\&E stained BCa tissue slides. This method can effectively overcome the drawbacks of patch-based methods by capturing the biological geometric structure of the cellular architecture at the entire WSI level. The proposed SlideGraph$^{+}$ can effectively incorporate both cell-level and contextual information by using different feature compositions and graph convolution. We also proposed a DAB density regression model which can predict HER2 specific DAB density directly from H\&E images. Experimental results for clinically important tasks of HER2 status prediction show that the proposed SlideGraph$^{+}$ method with estimated DAB density feature can produce higher accuracy than the state-of-the-art techniques. SlideGraph$^{+}$ can also be applied to other problems in computational pathology, such as recurrence and survival prediction, anti-HER2 treatment efficacy prediction.


\section*{Declaration of Competing Interest}
The authors confirm that there are no conflicts of interest.

\section*{Acknowledgments}
We would like to gratefully acknowledge all subjects whose whole slide image data has been used in the development of this study. This paper is supported by the PathLAKE Centre of Excellence for digital pathology and artificial intelligence which is funded from the Data to Early Diagnosis and Precision Medicine strand of the government’s Industrial Strategy Challenge Fund, managed and delivered by Innovate UK on behalf of UK Research and Innovation (UKRI). PathLAKE is a Research Ethics Committee (REC) approved research database, reference 19/SC/0363.

\section*{Author contributions statement}
WL: Experiment design and execution, methods development, analysis and write-up. MT and ER: Data provision (Nottingham Dataset) and clinical support. NR: Write-up, experiment design, supervision. FM: Conceptualisation,  method development, experiment design, analysis, writeup, supervision. 

\bibliographystyle{model2-names.bst}\biboptions{authoryear}
\bibliography{refs}

\begin{thebibliography}{51}
\expandafter\ifx\csname natexlab\endcsname\relax\def\natexlab#1{#1}\fi
\providecommand{\url}[1]{\texttt{#1}}
\providecommand{\href}[2]{#2}
\providecommand{\path}[1]{#1}
\providecommand{\DOIprefix}{doi:}
\providecommand{\ArXivprefix}{arXiv:}
\providecommand{\URLprefix}{URL: }
\providecommand{\Pubmedprefix}{pmid:}
\providecommand{\doi}[1]{\href{http://dx.doi.org/#1}{\path{#1}}}
\providecommand{\Pubmed}[1]{\href{pmid:#1}{\path{#1}}}
\providecommand{\bibinfo}[2]{#2}
\ifx\xfnm\relax \def\xfnm[#1]{\unskip,\space#1}\fi
\bibitem[{Acs et~al.(2020)Acs, Rantalainen and Hartman}]{acs2020artificial}
\bibinfo{author}{Acs, B.}, \bibinfo{author}{Rantalainen, M.},
  \bibinfo{author}{Hartman, J.}, \bibinfo{year}{2020}.
\newblock \bibinfo{title}{Artificial intelligence as the next step towards
  precision pathology}.
\newblock \bibinfo{journal}{Journal of internal medicine}
  \bibinfo{volume}{288}, \bibinfo{pages}{62--81}.
\bibitem[{Ahmad(2019)}]{ahmad2019breast}
\bibinfo{author}{Ahmad, A.}, \bibinfo{year}{2019}.
\newblock \bibinfo{title}{Breast Cancer Metastasis and Drug Resistance:
  Challenges and Progress}. volume \bibinfo{volume}{1152}.
\newblock \bibinfo{publisher}{Springer}.
\bibitem[{Ali et~al.(2013)Ali, Veltri, Epstein, Christudass and
  Madabhushi}]{ali2013cell}
\bibinfo{author}{Ali, S.}, \bibinfo{author}{Veltri, R.},
  \bibinfo{author}{Epstein, J.A.}, \bibinfo{author}{Christudass, C.},
  \bibinfo{author}{Madabhushi, A.}, \bibinfo{year}{2013}.
\newblock \bibinfo{title}{Cell cluster graph for prediction of biochemical
  recurrence in prostate cancer patients from tissue microarrays}, in:
  \bibinfo{booktitle}{Medical Imaging 2013: Digital Pathology},
  \bibinfo{organization}{International Society for Optics and Photonics}. p.
  \bibinfo{pages}{86760H}.
\bibitem[{Andrews et~al.(2002)Andrews, Hofmann and
  Tsochantaridis}]{andrews2002multiple}
\bibinfo{author}{Andrews, S.}, \bibinfo{author}{Hofmann, T.},
  \bibinfo{author}{Tsochantaridis, I.}, \bibinfo{year}{2002}.
\newblock \bibinfo{title}{Multiple instance learning with generalized support
  vector machines}, in: \bibinfo{booktitle}{AAAI/IAAI}, pp.
  \bibinfo{pages}{943--944}.
\bibitem[{Bandi et~al.(2018)Bandi, Geessink, Manson, Van~Dijk, Balkenhol,
  Hermsen, Bejnordi, Lee, Paeng, Zhong et~al.}]{bandi2018detection}
\bibinfo{author}{Bandi, P.}, \bibinfo{author}{Geessink, O.},
  \bibinfo{author}{Manson, Q.}, \bibinfo{author}{Van~Dijk, M.},
  \bibinfo{author}{Balkenhol, M.}, \bibinfo{author}{Hermsen, M.},
  \bibinfo{author}{Bejnordi, B.E.}, \bibinfo{author}{Lee, B.},
  \bibinfo{author}{Paeng, K.}, \bibinfo{author}{Zhong, A.}, et~al.,
  \bibinfo{year}{2018}.
\newblock \bibinfo{title}{From detection of individual metastases to
  classification of lymph node status at the patient level: the camelyon17
  challenge}.
\newblock \bibinfo{journal}{IEEE transactions on medical imaging}
  \bibinfo{volume}{38}, \bibinfo{pages}{550--560}.
\bibitem[{Campanella et~al.(2019)Campanella, Hanna, Geneslaw, Miraflor, Silva,
  Busam, Brogi, Reuter, Klimstra and Fuchs}]{campanella2019clinical}
\bibinfo{author}{Campanella, G.}, \bibinfo{author}{Hanna, M.G.},
  \bibinfo{author}{Geneslaw, L.}, \bibinfo{author}{Miraflor, A.},
  \bibinfo{author}{Silva, V.W.K.}, \bibinfo{author}{Busam, K.J.},
  \bibinfo{author}{Brogi, E.}, \bibinfo{author}{Reuter, V.E.},
  \bibinfo{author}{Klimstra, D.S.}, \bibinfo{author}{Fuchs, T.J.},
  \bibinfo{year}{2019}.
\newblock \bibinfo{title}{Clinical-grade computational pathology using weakly
  supervised deep learning on whole slide images}.
\newblock \bibinfo{journal}{Nature medicine} \bibinfo{volume}{25},
  \bibinfo{pages}{1301--1309}.
\bibitem[{Chami et~al.(2020)Chami, Abu-El-Haija, Perozzi, R{\'e} and
  Murphy}]{chami2020machine}
\bibinfo{author}{Chami, I.}, \bibinfo{author}{Abu-El-Haija, S.},
  \bibinfo{author}{Perozzi, B.}, \bibinfo{author}{R{\'e}, C.},
  \bibinfo{author}{Murphy, K.}, \bibinfo{year}{2020}.
\newblock \bibinfo{title}{Machine learning on graphs: A model and comprehensive
  taxonomy}.
\newblock \bibinfo{journal}{arXiv preprint arXiv:2005.03675} .
\bibitem[{Chew(1989)}]{chew1989constrained}
\bibinfo{author}{Chew, L.P.}, \bibinfo{year}{1989}.
\newblock \bibinfo{title}{Constrained delaunay triangulations}.
\newblock \bibinfo{journal}{Algorithmica} \bibinfo{volume}{4},
  \bibinfo{pages}{97--108}.
\bibitem[{Cruz-Roa et~al.(2014)Cruz-Roa, Basavanhally, Gonz{\'a}lez, Gilmore,
  Feldman, Ganesan, Shih, Tomaszewski and Madabhushi}]{cruz2014automatic}
\bibinfo{author}{Cruz-Roa, A.}, \bibinfo{author}{Basavanhally, A.},
  \bibinfo{author}{Gonz{\'a}lez, F.}, \bibinfo{author}{Gilmore, H.},
  \bibinfo{author}{Feldman, M.}, \bibinfo{author}{Ganesan, S.},
  \bibinfo{author}{Shih, N.}, \bibinfo{author}{Tomaszewski, J.},
  \bibinfo{author}{Madabhushi, A.}, \bibinfo{year}{2014}.
\newblock \bibinfo{title}{Automatic detection of invasive ductal carcinoma in
  whole slide images with convolutional neural networks}, in:
  \bibinfo{booktitle}{Medical Imaging 2014: Digital Pathology},
  \bibinfo{organization}{International Society for Optics and Photonics}. p.
  \bibinfo{pages}{904103}.
\bibitem[{Davis and Goadrich(2006)}]{davis2006relationship}
\bibinfo{author}{Davis, J.}, \bibinfo{author}{Goadrich, M.},
  \bibinfo{year}{2006}.
\newblock \bibinfo{title}{The relationship between precision-recall and roc
  curves}, in: \bibinfo{booktitle}{Proceedings of the 23rd international
  conference on Machine learning}, pp. \bibinfo{pages}{233--240}.
\bibitem[{Demir et~al.(2005)Demir, Gultekin and Yener}]{demir2005augmented}
\bibinfo{author}{Demir, C.}, \bibinfo{author}{Gultekin, S.H.},
  \bibinfo{author}{Yener, B.}, \bibinfo{year}{2005}.
\newblock \bibinfo{title}{Augmented cell-graphs for automated cancer
  diagnosis}.
\newblock \bibinfo{journal}{Bioinformatics} \bibinfo{volume}{21},
  \bibinfo{pages}{ii7--ii12}.
\bibitem[{Deng et~al.(2009)Deng, Dong, Socher, Li, Li and
  Fei-Fei}]{deng2009imagenet}
\bibinfo{author}{Deng, J.}, \bibinfo{author}{Dong, W.},
  \bibinfo{author}{Socher, R.}, \bibinfo{author}{Li, L.J.},
  \bibinfo{author}{Li, K.}, \bibinfo{author}{Fei-Fei, L.},
  \bibinfo{year}{2009}.
\newblock \bibinfo{title}{Imagenet: A large-scale hierarchical image database},
  in: \bibinfo{booktitle}{2009 IEEE conference on computer vision and pattern
  recognition}, \bibinfo{organization}{Ieee}. pp. \bibinfo{pages}{248--255}.
\bibitem[{Farahmand et~al.(2021)Farahmand, Fernandez, Ahmed, Rimm, Chuang,
  Reisenbichler and Zarringhalam}]{farahmand2021deep}
\bibinfo{author}{Farahmand, S.}, \bibinfo{author}{Fernandez, A.I.},
  \bibinfo{author}{Ahmed, F.S.}, \bibinfo{author}{Rimm, D.L.},
  \bibinfo{author}{Chuang, J.H.}, \bibinfo{author}{Reisenbichler, E.},
  \bibinfo{author}{Zarringhalam, K.}, \bibinfo{year}{2021}.
\newblock \bibinfo{title}{Deep learning trained on hematoxylin and eosin tumor
  region of interest predicts her2 status and trastuzumab treatment response in
  her2+ breast cancer}.
\newblock \bibinfo{journal}{Modern Pathology} , \bibinfo{pages}{1--8}.
\bibitem[{Fey and Lenssen(2019)}]{fey2019fast}
\bibinfo{author}{Fey, M.}, \bibinfo{author}{Lenssen, J.E.},
  \bibinfo{year}{2019}.
\newblock \bibinfo{title}{Fast graph representation learning with pytorch
  geometric}.
\newblock \bibinfo{journal}{arXiv preprint arXiv:1903.02428} .
\bibitem[{Gamper et~al.(2019)Gamper, Koohbanani, Benet, Khuram and
  Rajpoot}]{gamper2019pannuke}
\bibinfo{author}{Gamper, J.}, \bibinfo{author}{Koohbanani, N.A.},
  \bibinfo{author}{Benet, K.}, \bibinfo{author}{Khuram, A.},
  \bibinfo{author}{Rajpoot, N.}, \bibinfo{year}{2019}.
\newblock \bibinfo{title}{Pannuke: an open pan-cancer histology dataset for
  nuclei instance segmentation and classification}, in:
  \bibinfo{booktitle}{European Congress on Digital Pathology},
  \bibinfo{organization}{Springer}. pp. \bibinfo{pages}{11--19}.
\bibitem[{Graham et~al.(2019)Graham, Vu, Raza, Azam, Tsang, Kwak and
  Rajpoot}]{graham2019hover}
\bibinfo{author}{Graham, S.}, \bibinfo{author}{Vu, Q.D.},
  \bibinfo{author}{Raza, S.E.A.}, \bibinfo{author}{Azam, A.},
  \bibinfo{author}{Tsang, Y.W.}, \bibinfo{author}{Kwak, J.T.},
  \bibinfo{author}{Rajpoot, N.}, \bibinfo{year}{2019}.
\newblock \bibinfo{title}{Hover-net: Simultaneous segmentation and
  classification of nuclei in multi-tissue histology images}.
\newblock \bibinfo{journal}{Medical Image Analysis} \bibinfo{volume}{58},
  \bibinfo{pages}{101563}.
\bibitem[{Gunduz et~al.(2004)Gunduz, Yener and Gultekin}]{gunduz2004cell}
\bibinfo{author}{Gunduz, C.}, \bibinfo{author}{Yener, B.},
  \bibinfo{author}{Gultekin, S.H.}, \bibinfo{year}{2004}.
\newblock \bibinfo{title}{The cell graphs of cancer}.
\newblock \bibinfo{journal}{Bioinformatics} \bibinfo{volume}{20},
  \bibinfo{pages}{i145--i151}.
\bibitem[{He et~al.(2016)He, Zhang, Ren and Sun}]{he2016deep}
\bibinfo{author}{He, K.}, \bibinfo{author}{Zhang, X.}, \bibinfo{author}{Ren,
  S.}, \bibinfo{author}{Sun, J.}, \bibinfo{year}{2016}.
\newblock \bibinfo{title}{Deep residual learning for image recognition}, in:
  \bibinfo{booktitle}{Proceedings of the IEEE conference on computer vision and
  pattern recognition}, pp. \bibinfo{pages}{770--778}.
\bibitem[{Hou et~al.(2016)Hou, Samaras, Kurc, Gao, Davis and
  Saltz}]{hou2016patch}
\bibinfo{author}{Hou, L.}, \bibinfo{author}{Samaras, D.},
  \bibinfo{author}{Kurc, T.M.}, \bibinfo{author}{Gao, Y.},
  \bibinfo{author}{Davis, J.E.}, \bibinfo{author}{Saltz, J.H.},
  \bibinfo{year}{2016}.
\newblock \bibinfo{title}{Patch-based convolutional neural network for whole
  slide tissue image classification}, in: \bibinfo{booktitle}{Proceedings of
  the IEEE conference on computer vision and pattern recognition}, pp.
  \bibinfo{pages}{2424--2433}.
\bibitem[{Janowczyk and Madabhushi(2016)}]{janowczyk2016deep}
\bibinfo{author}{Janowczyk, A.}, \bibinfo{author}{Madabhushi, A.},
  \bibinfo{year}{2016}.
\newblock \bibinfo{title}{Deep learning for digital pathology image analysis: A
  comprehensive tutorial with selected use cases}.
\newblock \bibinfo{journal}{Journal of pathology informatics}
  \bibinfo{volume}{7}.
\bibitem[{Jaume et~al.(2021)Jaume, Pati, Anklin, Foncubierta and
  Gabrani}]{jaume2021histocartography}
\bibinfo{author}{Jaume, G.}, \bibinfo{author}{Pati, P.},
  \bibinfo{author}{Anklin, V.}, \bibinfo{author}{Foncubierta, A.},
  \bibinfo{author}{Gabrani, M.}, \bibinfo{year}{2021}.
\newblock \bibinfo{title}{Histocartography: A toolkit for graph analytics in
  digital pathology}, in: \bibinfo{booktitle}{MICCAI Workshop on Computational
  Pathology}, \bibinfo{organization}{PMLR}. pp. \bibinfo{pages}{117--128}.
\bibitem[{Kather et~al.(2019a)Kather, Heij, Grabsch, Kooreman, Loeffler, Echle,
  Krause, Muti, Niehues, Sommer et~al.}]{kather2019pan}
\bibinfo{author}{Kather, J.N.}, \bibinfo{author}{Heij, L.R.},
  \bibinfo{author}{Grabsch, H.I.}, \bibinfo{author}{Kooreman, L.F.},
  \bibinfo{author}{Loeffler, C.}, \bibinfo{author}{Echle, A.},
  \bibinfo{author}{Krause, J.}, \bibinfo{author}{Muti, H.S.},
  \bibinfo{author}{Niehues, J.M.}, \bibinfo{author}{Sommer, K.A.}, et~al.,
  \bibinfo{year}{2019}a.
\newblock \bibinfo{title}{Pan-cancer image-based detection of clinically
  actionable genetic alterations}.
\newblock \bibinfo{journal}{bioRxiv} , \bibinfo{pages}{833756}.
\bibitem[{Kather et~al.(2019b)Kather, Pearson, Halama, J{\"a}ger, Krause,
  Loosen, Marx, Boor, Tacke, Neumann et~al.}]{kather2019deep}
\bibinfo{author}{Kather, J.N.}, \bibinfo{author}{Pearson, A.T.},
  \bibinfo{author}{Halama, N.}, \bibinfo{author}{J{\"a}ger, D.},
  \bibinfo{author}{Krause, J.}, \bibinfo{author}{Loosen, S.H.},
  \bibinfo{author}{Marx, A.}, \bibinfo{author}{Boor, P.},
  \bibinfo{author}{Tacke, F.}, \bibinfo{author}{Neumann, U.P.}, et~al.,
  \bibinfo{year}{2019}b.
\newblock \bibinfo{title}{Deep learning can predict microsatellite instability
  directly from histology in gastrointestinal cancer}.
\newblock \bibinfo{journal}{Nature medicine} \bibinfo{volume}{25},
  \bibinfo{pages}{1054--1056}.
\bibitem[{Kingma and Ba(2014)}]{kingma2014adam}
\bibinfo{author}{Kingma, D.P.}, \bibinfo{author}{Ba, J.}, \bibinfo{year}{2014}.
\newblock \bibinfo{title}{Adam: A method for stochastic optimization}.
\newblock \bibinfo{journal}{arXiv preprint arXiv:1412.6980} .
\bibitem[{LeCun et~al.(1998)LeCun, Bottou, Bengio and
  Haffner}]{lecun1998gradient}
\bibinfo{author}{LeCun, Y.}, \bibinfo{author}{Bottou, L.},
  \bibinfo{author}{Bengio, Y.}, \bibinfo{author}{Haffner, P.},
  \bibinfo{year}{1998}.
\newblock \bibinfo{title}{Gradient-based learning applied to document
  recognition}.
\newblock \bibinfo{journal}{Proceedings of the IEEE} \bibinfo{volume}{86},
  \bibinfo{pages}{2278--2324}.
\bibitem[{Lu et~al.(2018)Lu, Wang, Prasanna, Corredor, Sedor, Bera, Velcheti
  and Madabhushi}]{lu2018feature}
\bibinfo{author}{Lu, C.}, \bibinfo{author}{Wang, X.},
  \bibinfo{author}{Prasanna, P.}, \bibinfo{author}{Corredor, G.},
  \bibinfo{author}{Sedor, G.}, \bibinfo{author}{Bera, K.},
  \bibinfo{author}{Velcheti, V.}, \bibinfo{author}{Madabhushi, A.},
  \bibinfo{year}{2018}.
\newblock \bibinfo{title}{Feature driven local cell graph (fedeg): Predicting
  overall survival in early stage lung cancer}, in:
  \bibinfo{booktitle}{International Conference on Medical Image Computing and
  Computer-Assisted Intervention}, \bibinfo{organization}{Springer}. pp.
  \bibinfo{pages}{407--416}.
\bibitem[{Lu et~al.(2019a)Lu, Duan, Orive-Miguel, Herve and
  Styles}]{lu2019graph}
\bibinfo{author}{Lu, W.}, \bibinfo{author}{Duan, J.},
  \bibinfo{author}{Orive-Miguel, D.}, \bibinfo{author}{Herve, L.},
  \bibinfo{author}{Styles, I.B.}, \bibinfo{year}{2019}a.
\newblock \bibinfo{title}{Graph-and finite element-based total variation models
  for the inverse problem in diffuse optical tomography}.
\newblock \bibinfo{journal}{Biomedical optics express} \bibinfo{volume}{10},
  \bibinfo{pages}{2684--2707}.
\bibitem[{Lu et~al.(2019b)Lu, Duan, Veesa and Styles}]{lu2019new}
\bibinfo{author}{Lu, W.}, \bibinfo{author}{Duan, J.}, \bibinfo{author}{Veesa,
  J.D.}, \bibinfo{author}{Styles, I.B.}, \bibinfo{year}{2019}b.
\newblock \bibinfo{title}{New nonlocal forward model for diffuse optical
  tomography}.
\newblock \bibinfo{journal}{Biomedical Optics Express} \bibinfo{volume}{10},
  \bibinfo{pages}{6227--6241}.
\bibitem[{Lu et~al.(2020)Lu, Graham, Bilal, Rajpoot and
  Minhas}]{lu2020capturing}
\bibinfo{author}{Lu, W.}, \bibinfo{author}{Graham, S.}, \bibinfo{author}{Bilal,
  M.}, \bibinfo{author}{Rajpoot, N.}, \bibinfo{author}{Minhas, F.},
  \bibinfo{year}{2020}.
\newblock \bibinfo{title}{Capturing cellular topology in multi-gigapixel
  pathology images}, in: \bibinfo{booktitle}{Proceedings of the IEEE/CVF
  Conference on Computer Vision and Pattern Recognition Workshops}, pp.
  \bibinfo{pages}{260--261}.
\bibitem[{M{\"u}llner(2011)}]{mullner2011modern}
\bibinfo{author}{M{\"u}llner, D.}, \bibinfo{year}{2011}.
\newblock \bibinfo{title}{Modern hierarchical, agglomerative clustering
  algorithms}.
\newblock \bibinfo{journal}{arXiv preprint arXiv:1109.2378} .
\bibitem[{Nahta et~al.(2006)Nahta, Yu, Hung, Hortobagyi and
  Esteva}]{nahta2006mechanisms}
\bibinfo{author}{Nahta, R.}, \bibinfo{author}{Yu, D.}, \bibinfo{author}{Hung,
  M.C.}, \bibinfo{author}{Hortobagyi, G.N.}, \bibinfo{author}{Esteva, F.J.},
  \bibinfo{year}{2006}.
\newblock \bibinfo{title}{Mechanisms of disease: understanding resistance to
  her2-targeted therapy in human breast cancer}.
\newblock \bibinfo{journal}{Nature clinical practice Oncology}
  \bibinfo{volume}{3}, \bibinfo{pages}{269--280}.
\bibitem[{Network et~al.(2012)}]{cancer2012comprehensive}
\bibinfo{author}{Network, C.G.A.}, et~al., \bibinfo{year}{2012}.
\newblock \bibinfo{title}{Comprehensive molecular portraits of human breast
  tumours}.
\newblock \bibinfo{journal}{Nature} \bibinfo{volume}{490}, \bibinfo{pages}{61}.
\bibitem[{Nguyen et~al.(2009)Nguyen, Torresani, De~La~Torre and
  Rother}]{nguyen2009weakly}
\bibinfo{author}{Nguyen, M.H.}, \bibinfo{author}{Torresani, L.},
  \bibinfo{author}{De~La~Torre, F.}, \bibinfo{author}{Rother, C.},
  \bibinfo{year}{2009}.
\newblock \bibinfo{title}{Weakly supervised discriminative localization and
  classification: a joint learning process}, in: \bibinfo{booktitle}{2009 IEEE
  12th International Conference on Computer Vision},
  \bibinfo{organization}{IEEE}. pp. \bibinfo{pages}{1925--1932}.
\bibitem[{Paszke et~al.(2017)Paszke, Gross, Chintala, Chanan, Yang, DeVito,
  Lin, Desmaison, Antiga and Lerer}]{paszke2017automatic}
\bibinfo{author}{Paszke, A.}, \bibinfo{author}{Gross, S.},
  \bibinfo{author}{Chintala, S.}, \bibinfo{author}{Chanan, G.},
  \bibinfo{author}{Yang, E.}, \bibinfo{author}{DeVito, Z.},
  \bibinfo{author}{Lin, Z.}, \bibinfo{author}{Desmaison, A.},
  \bibinfo{author}{Antiga, L.}, \bibinfo{author}{Lerer, A.},
  \bibinfo{year}{2017}.
\newblock \bibinfo{title}{Automatic differentiation in pytorch} .
\bibitem[{Prewitt(1979)}]{prewitt1979graphs}
\bibinfo{author}{Prewitt, J.M.}, \bibinfo{year}{1979}.
\newblock \bibinfo{title}{Graphs and grammars for histology: An introduction},
  in: \bibinfo{booktitle}{Proceedings of the Annual Symposium on Computer
  Application in Medical Care}, \bibinfo{organization}{American Medical
  Informatics Association}. p.~\bibinfo{pages}{18}.
\bibitem[{Qaiser et~al.(2018)Qaiser, Mukherjee, Reddy~Pb, Munugoti, Tallam,
  Pitk{\"a}aho, Lehtim{\"a}ki, Naughton, Berseth, Pedraza
  et~al.}]{qaiser2018her}
\bibinfo{author}{Qaiser, T.}, \bibinfo{author}{Mukherjee, A.},
  \bibinfo{author}{Reddy~Pb, C.}, \bibinfo{author}{Munugoti, S.D.},
  \bibinfo{author}{Tallam, V.}, \bibinfo{author}{Pitk{\"a}aho, T.},
  \bibinfo{author}{Lehtim{\"a}ki, T.}, \bibinfo{author}{Naughton, T.},
  \bibinfo{author}{Berseth, M.}, \bibinfo{author}{Pedraza, A.}, et~al.,
  \bibinfo{year}{2018}.
\newblock \bibinfo{title}{Her 2 challenge contest: a detailed assessment of
  automated her 2 scoring algorithms in whole slide images of breast cancer
  tissues}.
\newblock \bibinfo{journal}{Histopathology} \bibinfo{volume}{72},
  \bibinfo{pages}{227--238}.
\bibitem[{Rawat et~al.(2020)Rawat, Ortega, Roy, Sha, Shibata, Ruderman and
  Agus}]{rawat2020deep}
\bibinfo{author}{Rawat, R.R.}, \bibinfo{author}{Ortega, I.},
  \bibinfo{author}{Roy, P.}, \bibinfo{author}{Sha, F.},
  \bibinfo{author}{Shibata, D.}, \bibinfo{author}{Ruderman, D.},
  \bibinfo{author}{Agus, D.B.}, \bibinfo{year}{2020}.
\newblock \bibinfo{title}{Deep learned tissue “fingerprints” classify
  breast cancers by er/pr/her2 status from h\&e images}.
\newblock \bibinfo{journal}{Scientific reports} \bibinfo{volume}{10},
  \bibinfo{pages}{1--13}.
\bibitem[{Ross et~al.(2009)Ross, Slodkowska, Symmans, Pusztai, Ravdin and
  Hortobagyi}]{ross2009her}
\bibinfo{author}{Ross, J.S.}, \bibinfo{author}{Slodkowska, E.A.},
  \bibinfo{author}{Symmans, W.F.}, \bibinfo{author}{Pusztai, L.},
  \bibinfo{author}{Ravdin, P.M.}, \bibinfo{author}{Hortobagyi, G.N.},
  \bibinfo{year}{2009}.
\newblock \bibinfo{title}{The her-2 receptor and breast cancer: ten years of
  targeted anti--her-2 therapy and personalized medicine}.
\newblock \bibinfo{journal}{The oncologist} \bibinfo{volume}{14},
  \bibinfo{pages}{320--368}.
\bibitem[{Sharma et~al.(2015)Sharma, Zerbe, Lohmann, Kayser, Hellwich and
  Hufnagl}]{sharma2015review}
\bibinfo{author}{Sharma, H.}, \bibinfo{author}{Zerbe, N.},
  \bibinfo{author}{Lohmann, S.}, \bibinfo{author}{Kayser, K.},
  \bibinfo{author}{Hellwich, O.}, \bibinfo{author}{Hufnagl, P.},
  \bibinfo{year}{2015}.
\newblock \bibinfo{title}{A review of graph-based methods for image analysis in
  digital histopathology}.
\newblock \bibinfo{journal}{Diagnostic pathology} \bibinfo{volume}{1}.
\bibitem[{Sirinukunwattana et~al.(2018)Sirinukunwattana, Snead, Epstein, Aftab,
  Mujeeb, Tsang, Cree and Rajpoot}]{sirinukunwattana2018novel}
\bibinfo{author}{Sirinukunwattana, K.}, \bibinfo{author}{Snead, D.},
  \bibinfo{author}{Epstein, D.}, \bibinfo{author}{Aftab, Z.},
  \bibinfo{author}{Mujeeb, I.}, \bibinfo{author}{Tsang, Y.W.},
  \bibinfo{author}{Cree, I.}, \bibinfo{author}{Rajpoot, N.},
  \bibinfo{year}{2018}.
\newblock \bibinfo{title}{Novel digital signatures of tissue phenotypes for
  predicting distant metastasis in colorectal cancer}.
\newblock \bibinfo{journal}{Scientific reports} \bibinfo{volume}{8},
  \bibinfo{pages}{1--13}.
\bibitem[{Slamon et~al.(1987)Slamon, Clark, Wong, Levin, Ullrich and
  McGuire}]{slamon1987human}
\bibinfo{author}{Slamon, D.J.}, \bibinfo{author}{Clark, G.M.},
  \bibinfo{author}{Wong, S.G.}, \bibinfo{author}{Levin, W.J.},
  \bibinfo{author}{Ullrich, A.}, \bibinfo{author}{McGuire, W.L.},
  \bibinfo{year}{1987}.
\newblock \bibinfo{title}{Human breast cancer: correlation of relapse and
  survival with amplification of the her-2/neu oncogene}.
\newblock \bibinfo{journal}{science} \bibinfo{volume}{235},
  \bibinfo{pages}{177--182}.
\bibitem[{Tizhoosh and Pantanowitz(2018)}]{tizhoosh2018artificial}
\bibinfo{author}{Tizhoosh, H.R.}, \bibinfo{author}{Pantanowitz, L.},
  \bibinfo{year}{2018}.
\newblock \bibinfo{title}{Artificial intelligence and digital pathology:
  challenges and opportunities}.
\newblock \bibinfo{journal}{Journal of pathology informatics}
  \bibinfo{volume}{9}.
\bibitem[{Vahadane et~al.(2016)Vahadane, Peng, Sethi, Albarqouni, Wang, Baust,
  Steiger, Schlitter, Esposito and Navab}]{vahadane2016structure}
\bibinfo{author}{Vahadane, A.}, \bibinfo{author}{Peng, T.},
  \bibinfo{author}{Sethi, A.}, \bibinfo{author}{Albarqouni, S.},
  \bibinfo{author}{Wang, L.}, \bibinfo{author}{Baust, M.},
  \bibinfo{author}{Steiger, K.}, \bibinfo{author}{Schlitter, A.M.},
  \bibinfo{author}{Esposito, I.}, \bibinfo{author}{Navab, N.},
  \bibinfo{year}{2016}.
\newblock \bibinfo{title}{Structure-preserving color normalization and sparse
  stain separation for histological images}.
\newblock \bibinfo{journal}{IEEE transactions on medical imaging}
  \bibinfo{volume}{35}, \bibinfo{pages}{1962--1971}.
\bibitem[{Wang et~al.(2019)Wang, Sun, Liu, Sarma, Bronstein and
  Solomon}]{wang2019dynamic}
\bibinfo{author}{Wang, Y.}, \bibinfo{author}{Sun, Y.}, \bibinfo{author}{Liu,
  Z.}, \bibinfo{author}{Sarma, S.E.}, \bibinfo{author}{Bronstein, M.M.},
  \bibinfo{author}{Solomon, J.M.}, \bibinfo{year}{2019}.
\newblock \bibinfo{title}{Dynamic graph cnn for learning on point clouds}.
\newblock \bibinfo{journal}{Acm Transactions On Graphics (tog)}
  \bibinfo{volume}{38}, \bibinfo{pages}{1--12}.
\bibitem[{Weyn et~al.(1999)Weyn, van~de Wouwer, Kumar-Singh, van Daele,
  Scheunders, Van~Marck and Jacob}]{weyn1999computer}
\bibinfo{author}{Weyn, B.}, \bibinfo{author}{van~de Wouwer, G.},
  \bibinfo{author}{Kumar-Singh, S.}, \bibinfo{author}{van Daele, A.},
  \bibinfo{author}{Scheunders, P.}, \bibinfo{author}{Van~Marck, E.},
  \bibinfo{author}{Jacob, W.}, \bibinfo{year}{1999}.
\newblock \bibinfo{title}{Computer-assisted differential diagnosis of malignant
  mesothelioma based on syntactic structure analysis}.
\newblock \bibinfo{journal}{Cytometry: The Journal of the International Society
  for Analytical Cytology} \bibinfo{volume}{35}, \bibinfo{pages}{23--29}.
\bibitem[{Whitney et~al.(2018)Whitney, Corredor, Janowczyk, Ganesan, Doyle,
  Tomaszewski, Feldman, Gilmore and Madabhushi}]{whitney2018quantitative}
\bibinfo{author}{Whitney, J.}, \bibinfo{author}{Corredor, G.},
  \bibinfo{author}{Janowczyk, A.}, \bibinfo{author}{Ganesan, S.},
  \bibinfo{author}{Doyle, S.}, \bibinfo{author}{Tomaszewski, J.},
  \bibinfo{author}{Feldman, M.}, \bibinfo{author}{Gilmore, H.},
  \bibinfo{author}{Madabhushi, A.}, \bibinfo{year}{2018}.
\newblock \bibinfo{title}{Quantitative nuclear histomorphometry predicts
  oncotype dx risk categories for early stage er+ breast cancer}.
\newblock \bibinfo{journal}{BMC cancer} \bibinfo{volume}{18},
  \bibinfo{pages}{610}.
\bibitem[{Wolff et~al.(2018)Wolff, Hammond, Allison, Harvey, Mangu, Bartlett,
  Bilous, Ellis, Fitzgibbons, Hanna et~al.}]{wolff2018human}
\bibinfo{author}{Wolff, A.C.}, \bibinfo{author}{Hammond, M.E.H.},
  \bibinfo{author}{Allison, K.H.}, \bibinfo{author}{Harvey, B.E.},
  \bibinfo{author}{Mangu, P.B.}, \bibinfo{author}{Bartlett, J.M.},
  \bibinfo{author}{Bilous, M.}, \bibinfo{author}{Ellis, I.O.},
  \bibinfo{author}{Fitzgibbons, P.}, \bibinfo{author}{Hanna, W.}, et~al.,
  \bibinfo{year}{2018}.
\newblock \bibinfo{title}{Human epidermal growth factor receptor 2 testing in
  breast cancer: American society of clinical oncology/college of american
  pathologists clinical practice guideline focused update}.
\newblock \bibinfo{journal}{Archives of pathology \& laboratory medicine}
  \bibinfo{volume}{142}, \bibinfo{pages}{1364--1382}.
\bibitem[{Yarden(2001)}]{yarden2001biology}
\bibinfo{author}{Yarden, Y.}, \bibinfo{year}{2001}.
\newblock \bibinfo{title}{Biology of her2 and its importance in breast cancer}.
\newblock \bibinfo{journal}{Oncology} \bibinfo{volume}{61},
  \bibinfo{pages}{1--13}.
\bibitem[{Yener(2016)}]{yener2016cell}
\bibinfo{author}{Yener, B.}, \bibinfo{year}{2016}.
\newblock \bibinfo{title}{Cell-graphs: image-driven modeling of
  structure-function relationship}.
\newblock \bibinfo{journal}{Communications of the ACM} \bibinfo{volume}{60},
  \bibinfo{pages}{74--84}.
\bibitem[{Zhang et~al.(2006)Zhang, Platt and Viola}]{zhang2006multiple}
\bibinfo{author}{Zhang, C.}, \bibinfo{author}{Platt, J.C.},
  \bibinfo{author}{Viola, P.A.}, \bibinfo{year}{2006}.
\newblock \bibinfo{title}{Multiple instance boosting for object detection}, in:
  \bibinfo{booktitle}{Advances in neural information processing systems}, pp.
  \bibinfo{pages}{1417--1424}.
\bibitem[{Zhang and Goldman(2002)}]{zhang2002dd}
\bibinfo{author}{Zhang, Q.}, \bibinfo{author}{Goldman, S.A.},
  \bibinfo{year}{2002}.
\newblock \bibinfo{title}{Em-dd: An improved multiple-instance learning
  technique}, in: \bibinfo{booktitle}{Advances in neural information processing
  systems}, pp. \bibinfo{pages}{1073--1080}.

\end{thebibliography}

\clearpage

\onecolumn
\section*{\centering \LARGE Supplementary Material}

\begin{table*}[h!]
\caption{Overview of 15 quantitatively measured shape descriptors for assessing cell morphology.}\label{feature_15_explain}
\resizebox{\textwidth}{!}{%
\begin{tabular}{ll}
\hline
Feature                     & Description                                                                                                                                                         \\\hline
area                        & Number of pixels in the region                                                                                                                                      \\
bbox\_area                  & Number of pixels of bounding box of the region                                                                                                                      \\
convex\_area                & \begin{tabular}[c]{@{}l@{}}Number of pixels of convex hull image, which is the smallest convex polygon that \\ encloses the region\end{tabular}                     \\
eccentricity                & Ratio of the focal distance (distance between focal points) over the major axis length                                                                              \\
equivalent\_diameter        & The diameter of a circle with the same area as the region                                                                                                           \\
euler\_number               & Number of objects in the region subtracted from the number of holes in those objects                                                                                \\
extent                      & The proportion of pixels in the bounding box that are also in the region                                                                                            \\
filler\_area                & Number of pixels of the region will all the holes filled in                                                                                                         \\
inertia\_tensor\_eigvals\_x & The eigenvalues of the inertia tensor among x-axis in decreasing order                                                                                              \\
inertia\_tensor\_eigvals\_y & The eigenvalues of the inertia tensor among y-axis in decreasing order                                                                                              \\
major\_axis\_length         & \begin{tabular}[c]{@{}l@{}}The length of the major axis of the ellipse that has the same normalized second central \\ moments as the region\end{tabular}            \\
minor\_axis\_length         & \begin{tabular}[c]{@{}l@{}}The length of the minor axis of the ellipse that has the same normalized second central \\ moments as the region\end{tabular}            \\
orientation                 & \begin{tabular}[c]{@{}l@{}}Angle between the x-axis and the major axis of the ellipse that has the same second moments \\ as the region\end{tabular}                \\
perimeter                   & \begin{tabular}[c]{@{}l@{}}Perimeter of object which approximates the contour as a line through the centers of border \\ pixels using a 4-connectivity\end{tabular} \\
solidity                    & Ratio of pixels in the region to pixels of the convex hull image\\ \hline
\end{tabular}}
\end{table*}

\begin{table*}[h!]
\caption{AUPR comparison of SlideGraph$^{+}$ and the state-of-the-art methods on datasets from multiple centres. For all the results shown, the training is always done using the TCGA-BRCA dataset.}\label{Results_comparision_AUPR}
\resizebox{\textwidth}{!}{%
\begin{tabular}{llll}
\hline
Method         & Feature (dimension)                  & Test set                                                         & AUROC (mean$\pm$std)           \\\hline
\citep{kather2019pan}     &   Shufflenet                                      & TCGA-BRCA                                                             &  -                             \\
\citep{rawat2020deep}    &  Fingerprints (512)                                   & TCGA-BRCA                                                             &   -                              \\\hline
Linear regression & Maximum DAB density estimates (1)               & TCGA-BRCA                                                             & 0.18                   \\
               & Majority DAB density estimates (1)                   & TCGA-BRCA                                                             & 0.19                            \\
               & Average DAB density estimates (1)                    & TCGA-BRCA                                                             & 0.20                            \\\hline
SlideGraph$^{+}$      & Nuclear composition (5)                           & TCGA-BRCA                                                             & 0.32$\pm$0.01 \\
(Cross validation on               & Cellular morphology (30)                         & TCGA-BRCA                                                             & 0.34$\pm$0.05 \\
 TCGA-BRCA dataset)              & Resnet50 (2048)                            & TCGA-BRCA                                                             & 0.34$\pm$0.07  \\
               & DAB density estimates (4)                           & TCGA-BRCA                                                             & \textbf{0.37$\pm$0.03}                         \\
               & Nuclear composition + Cellular morphology (35)            & TCGA-BRCA                                                             & 0.37$\pm$0.05  \\
               & Nuclear composition + Cellular morphology + DAB (39)         & TCGA-BRCA                                                             & 0.36$\pm$0.07  \\\cline{1-4} 
SlideGraph$^{+}$                & DAB density estimates (4)                             & \begin{tabular}[c]{@{}l@{}}HER2C (0 / 3+)\end{tabular}         & \textbf{0.82$\pm$0.03}$^{*}$                                  \\
(Independent validation      &        & \begin{tabular}[c]{@{}l@{}}HER2C (0, 1+ / 3+)\end{tabular}     & 0.64$\pm$0.05$^{*}$                               \\
 on HER2C and               &                & \begin{tabular}[c]{@{}l@{}}Nott-HER2 (0 / 3+)\end{tabular}     & \textbf{0.25$\pm$0.04}$^{*}$                                  \\
Nott-HER2 datasets)               &                  & \begin{tabular}[c]{@{}l@{}}Nott-HER2 (0, 1+ / 3+)\end{tabular} & 0.20$\pm$0.04$^{*}$                             \\
               &                   & \begin{tabular}[c]{@{}l@{}}Nott-HER2 (- / +)\end{tabular}      & 0.28$\pm$0.03$^{*}$              \\ \hline  

\end{tabular}}
{\small \raggedright *: AUPR in the independent test sets are not comparable because the number of cases in both classes and the ratio between them varies among different sets. \par}
\end{table*}
\clearpage

\end{document}